%% file: neurips_2026.tex
\documentclass{article}

\usepackage[preprint]{neurips_2026}
\usepackage[utf8]{inputenc} 
\usepackage[T1]{fontenc}    
\usepackage{hyperref}       
\usepackage{url}            
\usepackage{booktabs}       
\usepackage{amsfonts}       
\usepackage{nicefrac}       
\usepackage{microtype}      
\usepackage{xcolor}         

\usepackage{graphicx} 
\usepackage{amsmath}
\usepackage{algorithm}
\usepackage{algorithmic}
\usepackage{diagbox}
\usepackage{multirow}
\usepackage{subcaption}

\title{Cumulative Meta-Learning from Active Learning Queries for Robustness to Spurious Correlations}

\author{
    Chew Kin Whye \quad Wang Jingxian \\
    Department of Computer Science \\
    National University of Singapore \\
    Singapore, S119077 \\
    \texttt{\{kinwhye,wang\}@nus.edu.sg}
}

\begin{document}

\maketitle

\input{Sections/A_Abstract}
\input{Sections/B_Introduction}

\input{Sections/D_PriorWorks}
\input{Sections/E_ProposedMethod}
\input{Sections/F_Experiments}
\input{Sections/G_Limitations}
\input{Sections/H_Conclusion}

\bibliographystyle{plainnat}
\bibliography{references}
\appendix
\newpage
\input{Appendix/A_Method}

\newpage
\input{Appendix/B_TheoreticalProof}
\newpage
\input{Appendix/C_Datasets}
\newpage
\input{Appendix/D_ExperimentalSetup}
\newpage
\input{Appendix/E_Computational_Resources}

\newpage
\input{Appendix/F_ExperimentResults}

\end{document}

%% file: Sections/A_Abstract.tex
\begin{abstract}

Spurious correlations in real-world datasets pose significant challenges for machine learning models, leading to reliance on irrelevant patterns that compromise reliability, generalization, and fairness. Active learning offers a promising way to mitigate this failure mode by querying informative samples that distinguish between core and spurious correlations. However, standard active-learning frameworks simply append queried examples to the labeled set, effectively updating only the likelihood term, which has a limited effect in deep learning regimes where informative samples are diluted by the much larger labeled set and can be easily memorized by overparameterized models. We propose Cumulative Active Meta-Learning (CAML), a novel active-learning framework that uses actively queried examples to meta-learn the prior, or inductive bias, governing how the model adapts, a strategy that remains effective because meta-learning amplifies their impact and promotes generalization. CAML casts each active-learning round as a meta-learning task: the current labeled set serves as meta-train data for adaptation, the newly queried batch serves as meta-test data to evaluate generalization, while CAML meta-learns the inductive bias that promotes this generalization. Crucially, unlike conventional meta-learning, which treats meta-learning tasks as independently and identically distributed (i.i.d.), CAML exploits the sequential dependence between active-learning rounds by maintaining a cumulative inductive bias that is progressively refined. Theoretically, we show that this cumulative formulation introduces interaction terms that couple earlier meta-learned inductive biases with later query-induced objectives, capturing dependencies absent from standard meta-learning. Empirically, CAML improves minority-group accuracy across popular spurious-correlation benchmarks and acquisition strategies, with gains of up to 27.8\% on Dominoes, 29.9\% on Waterbirds, 14.3\% on SpuCo, and 24.0\% on CivilComments.
\end{abstract}

%% file: Sections/B_Introduction.tex
\section{Introduction}
\label{section:introduction}
Real-world datasets often contain \textit{spurious correlations}: patterns that are predictive of labels in the training data but irrelevant to the \textit{true target function}~\citep{https://doi.org/10.48550/arxiv.2204.02937}. A canonical example is the Waterbirds dataset~\citep{DBLP:journals/corr/abs-1911-08731}, where the task is to classify bird species as waterbird or landbird, which is spuriously correlated with the water and land backgrounds. \citet{damour2020underspecification} propose that the training dataset is \textit{underspecified}, where multiple plausible hypotheses describe the data equally well, with no additional evidence to prefer one over another. Consequently, training the model via \textit{Empirical Risk Minimization (ERM)} will indiscriminately absorb all correlations, both core and spurious, present in the training data~\citep{arjovsky2020invariant}, exploiting them to minimize the empirical risk. These models do not accurately capture the underlying target function, leading to poor generalization, reduced reliability, and unfairness.

One natural way to reduce this underspecification is to acquire additional labels on inputs that distinguish between competing hypotheses. This is precisely the promise of \textit{active learning}, strategically selecting the most informative unlabeled examples for annotation to maximize performance under a limited labeling budget~\citep{settles.tr09}. In the context of spurious correlations, \citet{tamkin2022active} found that existing uncertainty-based acquisition functions can automatically resolve this underspecification by selecting data points where the core and spurious correlations disagree, which are the data points most informative for distinguishing between them. However, in standard active learning pipelines, these informative queries are simply appended to the labeled set and used for supervised training, thereby updating only the \textit{likelihood distribution}, i.e., \emph{what the model adapts to}. This underutilizes queried samples in modern deep learning regimes, where the signal from small but highly informative query batches can be diluted by much larger labeled datasets or memorized by overparameterized models without meaningfully changing the underlying classifier.

To address this limitation, we present \textit{Cumulative Active Meta-Learning} (CAML), a framework that amplifies the impact of actively queried samples by using them not only to update the model's fit but also to update its \textit{inductive bias} governing \textit{how the model adapts}. CAML casts each active learning round as a meta-learning task: the currently labeled dataset serves as the meta-train set for learning correlations, the newly queried batch serves as the meta-test set to evaluate generalizability, and meta-learning updates the inductive bias that promotes this generalization. This remains effective in modern deep learning regimes because meta-learning keeps the queried batch as a distinct meta-test signal that shapes the inductive bias governing adaptation, thereby amplifying its influence rather than diluting it into the larger labeled set and encouraging generalization rather than memorization.

\begin{figure*}[t]
    \centering
    \begin{subfigure}[t]{0.32\textwidth}
        \centering
        \includegraphics[width=\linewidth]{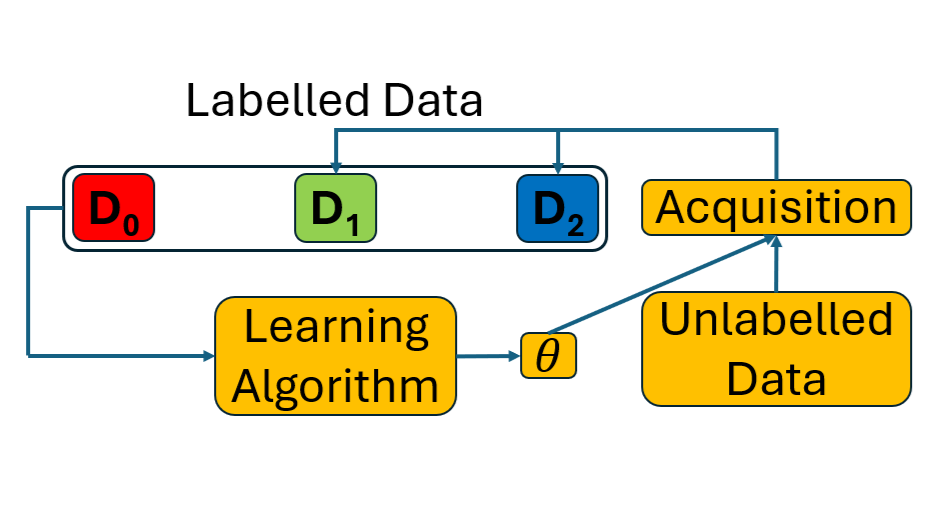}
        \caption{Active Learning}
    \end{subfigure}
    \hfill
    \begin{subfigure}[t]{0.32\textwidth}
        \centering
        \includegraphics[width=\linewidth]{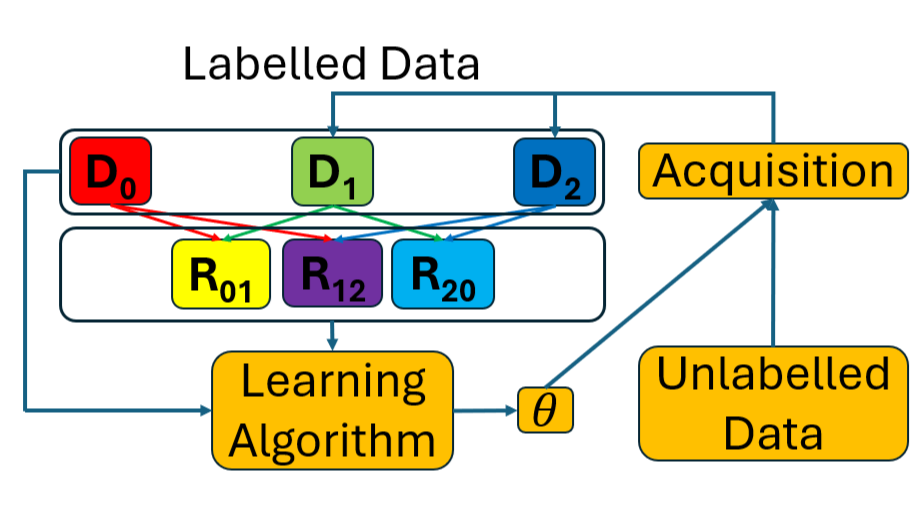}
        \caption{Meta-Learning}
    \end{subfigure}
    \hfill
    \begin{subfigure}[t]{0.32\textwidth}
        \centering
        \includegraphics[width=\linewidth]{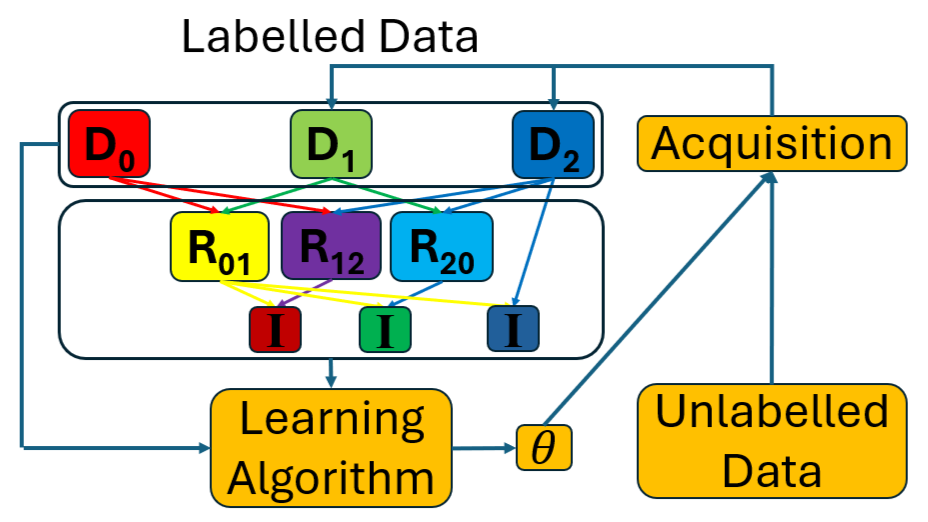}
        \caption{CAML}
    \end{subfigure}
    \caption{Conceptual comparison of standard active learning, conventional meta-learning, and CAML. Standard active learning simply appends informative queried samples to the labeled set. Conventional meta-learning uses queries to learn independent pairwise inductive-bias terms, $R_{ij}$.
    CAML yields additional interaction terms $\mathcal{I}$ that couple earlier inductive biases with later query-induced objectives.}
    \label{fig:illustration}
\end{figure*}

A key challenge, however, is that active learning induces a sequentially dependent learning process, in which each query batch is informative relative to the inductive bias accumulated through earlier rounds. Conventional meta-learning~\citep{finn2017modelagnosticmetalearningfastadaptation, 9428530}, however, treats meta-learning tasks as independently and identically distributed (i.i.d.), so applying it independently across active-learning rounds learns the inductive bias from each queried batch in isolation, without accounting for the inductive biases accumulated over earlier rounds. CAML addresses this limitation through a novel meta-learning algorithm that learns a \textit{cumulative inductive bias} across active-learning rounds, conditioning each newly queried batch on this cumulated bias to learn the incremental update needed to generalize to that batch. By doing so, CAML progressively refines the inductive bias across active-learning rounds towards core features and away from spurious correlations, yielding more label-efficient generalization performance. Theoretically, we show that CAML captures the sequential dependence of active learning by introducing additional interaction terms that couple previously learned inductive biases with later query-induced objectives. Figure~\ref{fig:illustration} provides an overview of the conceptual progression from standard active learning and conventional meta-learning to CAML.

Our work presents the following key contributions:

\begin{enumerate}
    \item \textbf{Meta-Learning \textit{Framework} for Active Learning:}
    We propose a novel framework for active learning in which informative, actively queried samples are used to learn the prior/inductive bias governing how the model adapts, thereby amplifying their impact in deep learning regimes. Specifically, CAML casts active learning as a sequence of meta-learning tasks, where the previously labeled data serves as the meta-train set, the newly queried samples serve as the meta-test set, and meta-learning optimizes the inductive bias for generalization. 
        
    \item \textbf{Meta-Learning \textit{Algorithm} for Sequentially Dependent Active Learning:} We propose a novel meta-learning algorithm tailored to the sequential nature of active learning, which progressively refines the cumulative inductive bias across rounds. Specifically, CAML conditions each new task on the cumulative inductive bias learned from previous tasks and learns the incremental update required for generalization.
        
    \item \textbf{Theoretical Justification and Empirical Validation:} We show theoretically that CAML captures dependencies absent from conventional meta-learning through interaction terms that couple previously learned inductive biases with later query-induced objectives. We further validate CAML across multiple spurious-correlation benchmarks and acquisition strategies, demonstrating consistent gains in minority-group accuracy and thereby improved robustness to spurious correlations.
\end{enumerate}

%% file: Sections/D_PriorWorks.tex
\section{Related Works}
\label{section:prior}
\subsection{Spurious Correlation}
\label{section:prior_spurious}
One line of work assumes access only to the \textbf{original biased training set}, aiming either to infer spurious attributes implicitly and use them to train a more robust classifier~\citep{nam2020learning,utama2020debiasing,dagaev2021toogoodtobetrue,DBLP:journals/corr/abs-2107-09044,yaghoobzadeh2021increasing,https://doi.org/10.48550/arxiv.2203.01517} or emphasize learning a diverse set of features to prevent the model from over-relying on the simplest features~\citep{pagliardini2022agree, taghanaki2022masktune,lee2023diversify}. A second line assumes access to \textbf{multiple environments}~\citep{arjovsky2020invariant, pmlr-v119-ahuja20a, Lin_2022_CVPR, pmlr-v162-zhou22e} and learns predictors that exhibit invariance across them, based on the premise that spurious correlations vary across environments while the core features remain invariant. A third line assumes access to \textbf{spurious attribute annotations} and uses them to rebalance training through methods such as Subsampling, Reweighting, and Group Distributionally Robust Optimization (gDRO)~\citep{DBLP:journals/corr/abs-1911-08731}, such that the spurious attribute is no longer sufficiently predictive of the label under the training objective. Relatedly, Deep Feature Reweighting~\citep{https://doi.org/10.48550/arxiv.2204.02937} uses a \textbf{balanced target dataset}, in which the spurious correlation no longer holds, to retrain the classification head of an ERM model.

\citet{tamkin2022active} studies a different regime: \textbf{active learning} as a sample-efficient strategy for automatically mitigating spurious correlations, without requiring privileged information, such as spurious-attribute annotations, environment identifiers, or balanced target datasets. They find that existing uncertainty-based algorithms succeed by selecting examples on which the core and spurious correlations disagree, making these examples especially informative for identifying the intended target correlation. Our paper extends the work of~\citet{tamkin2022active} by proposing a novel meta-learning active learning framework and algorithm that more effectively leverages actively queried data points to resolve spurious correlations by learning an inductive bias towards the generalizable core features.

\subsection{Active Learning}
Active learning aims to improve label efficiency by selecting the most informative unlabeled examples for annotation. Most prior work predominantly centers on the design of \textit{acquisition functions}, which dictate the selection of data points to be labeled. These acquisition functions are typically based on two main heuristics: \textit{uncertainty} and \textit{representativeness}. \textbf{Uncertainty-based} methods prioritize examples on which the current model is most uncertain, aiming to maximally reduce the model's uncertainty~\citep{ash2020deep}. This uncertainty can be estimated from the model's predictive probabilities~\citep{10.1007/11871842_40, 6889457, ducoffe2018adversarial}, or from the disagreement or variance across an ensemble of models~\citep{houlsby2011bayesian, gal2017deep, 8579074}. \textbf{Representative-based} methods instead prioritize examples that best reflect the entire unlabeled dataset, thereby serving as a surrogate for the broader data distribution~\citep{ash2020deep}. Such representativeness can be encouraged using CoreSet approaches~\citep{phillips2016coresets, geifman2017deep, sener2018active} or discriminator-based methods~\citep{gissin2019discriminative}. \textbf{Hybrid} approaches combine these heuristics to leverage the strengths of both, either by adaptively selecting among existing acquisition strategies based on their estimated utility~\citep{Baram2003OnlineCO, Hsu_Lin_2015} or by directly constructing metrics that capture both properties~\citep{NIPS2010_5487315b, ash2020deep, pmlr-v108-shui20a}. We refer the reader to \citet{settles.tr09} and \citet{ren2021survey} for comprehensive surveys.

Our paper is complementary to the active learning literature on acquisition design, focusing on how the actively queried data points should be integrated to amplify their impact. Instead of treating queried samples as ordinary extensions of the labeled set that only refine the likelihood, we integrate them through meta-learning to update the prior, which is more effective in deep learning regimes.

\subsection{Meta-Learning}
We adopt the bilevel optimization view of meta-learning~\citep{9428530}, in which outer-level meta-parameters are optimized so that, after inner-level adaptation on a meta-train dataset, the resulting model performs well on a meta-test dataset. From this perspective, meta-learning can be understood as learning the prior, or inductive bias, of the learning algorithm that governs how a model adapts to data, which could represent various aspects such as the parameter initialization~\citep{finn2017modelagnosticmetalearningfastadaptation, nichol2018firstordermetalearningalgorithms, rusu2019metalearninglatentembeddingoptimization}, optimizer~\citep{andrychowicz2016learninglearngradientdescent, duan2016rl2fastreinforcementlearning, li2017metasgdlearninglearnquickly}, loss function~\citep{santos2017learninglossfunctionssemisupervised, wu2018learningteachdynamicloss, ren2019learningreweightexamplesrobust}, or the model architecture~\citep{zoph2017neuralarchitecturesearchreinforcement, liu2019dartsdifferentiablearchitecturesearch, real2020automlzeroevolvingmachinelearning}.

Meta-learning has also been studied in settings related to active learning and spurious-correlation mitigation. In active learning, prior work has used meta-learning to learn acquisition policies~\citep{pang2018metalearningtransferableactivelearning, schrum2020metaactivelearningprobabilisticallysafeoptimization}, initialize active learners for faster adaptation to newly queried examples~\citep{zhu-etal-2022-shot, ho2024learninglearnfewshotcontinual}, and estimate uncertainty for query selection~\citep{NEURIPS2018_8e2c381d, kim2018bayesianmodelagnosticmetalearning}. In spurious correlation mitigation, \citet{zheng2024spuriousnessawaremetalearninglearningrobust} applies meta-learning in a different setup where spurious attributes are identified using a pre-trained vision-language model.

Our work differs in two key respects. First, we use meta-learning to exploit queried samples after acquisition by learning an inductive bias that promotes generalization. Second, unlike conventional meta-learning, which treats tasks as i.i.d. and learns an independent inductive bias for each task, our method accounts for the sequential dependence inherent in active learning and, correspondingly, refines the cumulatively learned inductive bias across rounds.

%% file: Sections/E_ProposedMethod.tex
\section{Cumulative Active Meta-Learning (CAML)}
\label{section:CAML}
We now present Cumulative Active Meta-Learning (CAML), our novel framework and algorithm for leveraging actively queried samples to learn the prior, rather than simply appending them to the training set to update the likelihood.
We first motivate this view by showing why likelihood-only updates can underutilize queried samples in modern deep learning regimes, and why updating the prior offers a more effective way to exploit their information.  We then instantiate this prior-learning perspective through meta-learning, where queried samples serve as meta-test signals that shape how the model adapts to the currently labeled data.  Finally, we show that active learning induces sequentially dependent learning problems, motivating CAML as a cumulative meta-learning objective that captures this dependency by progressively refining the inductive bias across rounds. Theoretically, we show that this cumulative formulation introduces interaction terms that couple earlier query-induced inductive biases with later objectives, capturing dependencies absent from conventional meta-learning.

\subsection{Likelihood vs Prior in Active Learning}
\label{sec:theory-motivation}

To clarify how actively queried samples influence learning, we view supervised training through the lens of \textit{Maximum A Posteriori} (MAP) estimation. Given a labeled dataset $\mathcal{D}$, the learner seeks parameters that maximize the \textit{training objective} that is represented by the posterior probability:

\begin{equation}
\label{eqn:posterior}
\theta^{\star}
= \arg\max_{\theta} p(\theta \mid \mathcal{D})
= \arg\max_{\theta} \frac{p(\mathcal{D}\mid \theta)p(\theta)}{p(\mathcal{D})}
= \arg\max_{\theta} p(\mathcal{D}\mid \theta)p(\theta).
\end{equation}
Here, $p(\theta \mid \mathcal{D})$ denotes the \textit{posterior} over model parameters, $p(\mathcal{D}\mid\theta)$ is the \textit{likelihood} and $p(\theta)$ is the \textit{prior}. Equation~\ref{eqn:posterior} provides a useful conceptual decomposition of the training objective. The likelihood term captures how well the model fits the observed labeled data, while the prior encodes preferences over solutions, i.e., the inductive bias that shapes how the model generalizes beyond the training set.

In standard active learning, the learner queries a batch of unlabeled examples $\mathcal{D}_q$ for annotation to maximize the model's performance. After labeling, however, $\mathcal{D}_q$ is typically added to the original training dataset $\mathcal{D}_o$. Under the MAP decomposition above, the \emph{posterior is updated solely through the likelihood term}, while the prior remains untouched. In practice, this corresponds to minimizing the standard \textit{supervised loss} (e.g., cross-entropy, mean-squared error) on the expanded dataset, thereby shrinking the set of parameters consistent with the dataset and reducing uncertainty in the likelihood term. However, we argue that this is insufficient in typical deep learning contexts for two key reasons:

\begin{enumerate}
    \item \textbf{Dilution by large labeled datasets.} 
    In active learning, the queried batch $D_q$ is typically small under limited annotation budgets, as it aims to maximize model performance with few additional labels~\citep{settles.tr09}. In contrast, modern deep learning typically operates on already large labeled datasets $D_o$ required for competitive performance~\citep{LeCun2015}. Under the standard pipeline, the newly labeled batch $D_q$ is simply merged into $D_o$. The likelihood term, instantiated by the supervised loss, is then evaluated on the expanded dataset $D = D_o \cup D_q$:
    \begin{equation}
    \notag
    \mathcal{L}_{\mathcal{D}}(\theta) =
    \frac{|\mathcal{D}_q|}{|\mathcal{D}_o \cup \mathcal{D}_q|} \mathcal{L}_{\mathcal{D}_q}(\theta) + \frac{|\mathcal{D}_o|}{|\mathcal{D}_o \cup \mathcal{D}_q|} \mathcal{L}_{\mathcal{D}_o}(\theta).
    \end{equation}    
    Thus, since $|\mathcal{D}_q| \ll |\mathcal{D}_o|$, the supervised loss is dominated by $D_o$, causing the \emph{signal from even highly informative queried examples to be diluted}, thereby limiting their impact.
    
    \item \textbf{Memorization by overparameterized deep networks.} Modern deep learning often operates in an overparameterized regime, where model capacity is often increased beyond interpolation to yield improved test performance~\citep{belkin2019reconciling, nakkiran2020deep}. In this regime, such networks have sufficient capacity to memorize arbitrary training labels~\citep{zhang2017understandingdeeplearningrequires}, and can easily \emph{memorize $D_q$ while leaving the original classifier essentially unchanged}. The effect is especially pronounced under spurious correlations, where the model learns the spurious feature for the majority group and simply memorizes the small minority-group counterexamples instead of learning the robust signal~\citep{pmlr-v119-sagawa20a, bayat2024pitfallsmemorizationmemorizationhurts, You2025}.
\end{enumerate}

Our framework seeks an orthogonal route towards using the queried data points $D_q$ to shape the posterior $p(\theta\mid D)$ by learning the prior $p(\theta)$. The prior expresses a preference over solutions and therefore encodes the learner's \textit{inductive bias}: the built-in assumptions of a learning algorithm that shape how the model adapts to the observed data and extrapolates beyond it, a condition central to generalization~\citep{10.1162/neco.1996.8.7.1341, goyal2022inductive}. This addresses the two limitations above:

\begin{enumerate}
    \item \textbf{Amplification through inductive bias}: The queried batch $D_q$ is not diluted with the original labeled set $D_o$, remaining as separate supervision signals for different parameters: $D_o$ drives the standard model update of $\theta$, whereas $D_q$ drives the meta-update. More importantly, this amplifies their impact since $D_q$ updates the inductive bias that governs how the model adapts on $D_o$~\citep{baxter1998theoretical, grant2018recastinggradientbasedmetalearninghierarchical}, such as in~\citep{li2017metasgdlearninglearnquickly, ren2019learningreweightexamplesrobust, real2020automlzeroevolvingmachinelearning}.

    \item \textbf{Optimization for generalization}: Meta-learning uses $D_o$ as the meta-train set and $D_q$ as the meta-test set, and optimizes the inductive bias so that performance transfers from adaptation on $D_o$ to evaluation on $D_q$. Because generalization is precisely such a transfer across datasets, the objective is directly aligned with extracting signals from $D_q$ to shape the inductive bias that promotes generalization~\citep{nichol2018firstordermetalearningalgorithms}.
\end{enumerate}

\subsection{Cumulative Active Meta-Learning}
\label{sec:CAML}
We now instantiate the prior-learning view from Section~\ref{sec:theory-motivation} using meta-learning. In each active learning round, we treat the current labeled set $D_o$ as the meta-train set and the newly queried batch $D_q$ as the meta-test set. The objective is to learn an inductive bias under which adaptation on $D_o$ generalizes to $D_q$. Using a one-step MAML instantiation, together with its interpretation as a meta-learning regularizer~\citep{li2017learninggeneralizemetalearningdomain}, this yields

\begin{equation}
    \label{eqn:MLDG}
    L(\theta) = \mathcal{L}_{\mathcal{D}_o}(\theta) + \mathcal{L}_{\mathcal{D}_q}\bigl(\theta - \eta_{in} \nabla_{\theta}\mathcal{L}_{\mathcal{D}_o}(\theta)\bigr).
\end{equation}

The first term $\mathcal{L}_{D_o}(\theta)$ is the supervised loss on the current labeled dataset $D_o$.
The second term $\mathcal{L}_{D_q}\bigl(\theta - \eta_{in} \nabla_{\theta}\mathcal{L}_{D_o}(\theta)\bigr)$ evaluates the supervised loss on the queried batch $D_q$ after a one-step update on $D_o$. This can be interpreted as the \textit{meta-learning regularization term} that encodes the inductive bias ensuring the model generalizes to $D_q$ after adaptation on $D_o$.

In active learning, however, this process unfolds over multiple rounds rather than a single meta-learning episode. Starting from an initial labeled set $\mathcal{D}_0$, the learner acquires a queried batch $\mathcal{D}_i$ at each round $i \in \{1,\dots,n\}$, yielding a sequence $\mathcal{D}_1,\ldots,\mathcal{D}_n$. Importantly, the batch acquired at round $i$ is selected by an acquisition function applied to the model trained under the objective from the previous round $L_{i-1}$. Since $L_{i-1}$ already encodes the inductive bias learned from earlier rounds, the resulting query batch $D_i$ is informative relative to that cumulated bias. Consequently, the meta-learning tasks induced by active learning are not independently and identically distributed (i.i.d.) but inherently sequentially dependent, and suggest that the task induced at round $i$ should be formulated as an \emph{incremental update to the previously learned cumulative inductive biases}, rather than as an independent one-round meta-learning task.

To capture this sequential dependence, we propose \textit{Cumulative Active Meta-Learning} (CAML), a novel meta-learning algorithm that progressively refines the inductive bias across active-learning rounds. CAML maintains a cumulative training objective across rounds and uses it to define the adaptation step for each newly queried batch. At round $n$, we define the training objective
\begin{equation}
\label{eqn:CAML}
L_n(\theta)=\mathcal{L}_{\mathcal{D}_0}(\theta)+\sum_{i=1}^{n}\mathcal{L}_{\mathcal{D}_i}\bigl(\theta-\eta_{\mathrm{in}}\nabla_{\theta}L_{i-1}(\theta)\bigr),
\end{equation}

where $\mathcal{L}_{\mathcal{D}_0}(\theta)$ is the supervised loss on the initial labeled set. The summation aggregates the meta-learning regularization terms from all active learning rounds. For each round $i$, $\mathcal{L}_{\mathcal{D}_i}\bigl(\theta-\eta_{\mathrm{in}}\nabla_{\theta}L_{i-1}(\theta)\bigr)$ denotes the corresponding meta-learning regularization term, which evaluates performance on the queried batch $\mathcal{D}_i$ after one inner update taken with respect to the previous cumulative objective $L_{i-1}$. By taking the inner update with respect to $L_{i-1}$, adaptation at round $i$ is conditioned on the cumulative inductive bias learned so far. The resulting meta-learning regularization term, therefore, encourages the \emph{incremental inductive bias required for generalization} to the newly queried batch. We provide an efficient forward-recursion implementation in Appendix~A, which requires $n+1$ forward and backward passes per optimization step. Additionally, we use the first-order MAML approximation~\citep{finn2017modelagnosticmetalearningfastadaptation, nichol2018firstordermetalearningalgorithms} to avoid any expensive second-order derivatives.

\subsection{Theoretical and Conceptual Analysis}
\label{section:theoretical_analysis}

We now theoretically analyze and justify the CAML objective in Equation~\ref{eqn:CAML} by contrasting it with conventional meta-learning. We show that standard meta-learning induces independent pairwise inductive bias terms, whereas CAML additionally introduces interaction terms that explicitly couple the inductive bias learned in earlier rounds with the objectives introduced later. To make this comparison tractable, we consider a simplified two-round active learning setting with an initial labeled dataset $\mathcal{D}_0$ and two queried batches $\mathcal{D}_1$ and $\mathcal{D}_2$. Following~\citet{li2017learninggeneralizemetalearningdomain, li2020sequentiallearningdomaingeneralization}, we analyze the resulting objectives using a first-order Taylor approximation. Section~\ref{sec:independent-meta} first derives the standard MAML approximation, and Section~\ref{sec:caml_theory} then derives the corresponding CAML approximation.

\subsubsection{Standard Meta-Learning in a Two-Round Setting}
\label{sec:independent-meta}
We first analyze the conventional one-step MAML, where each round induces an independent meta-learning task. In the two-round setting, this gives the three meta-learning tasks with meta-train/meta-test pairs $(D_0, D_1)$, $(D_1, D_2)$, and $(D_2, D_0)$, yielding the objective

\begin{equation}
\begin{aligned}
\label{eqn:lmaml}
L_{\mathrm{MAML}}(\theta)
&=
\mathcal{L}_{\mathcal{D}_0}(\theta)
+
\mathcal{L}_{\mathcal{D}_1}\bigl(\theta-\eta_{\mathrm{in}}\nabla_\theta \mathcal{L}_{\mathcal{D}_0}(\theta)\bigr) \\
&+
\mathcal{L}_{\mathcal{D}_1}(\theta)
+
\mathcal{L}_{\mathcal{D}_2}\bigl(\theta-\eta_{\mathrm{in}}\nabla_\theta \mathcal{L}_{\mathcal{D}_1}(\theta)\bigr) \\
&+
\mathcal{L}_{\mathcal{D}_2}(\theta)
+
\mathcal{L}_{\mathcal{D}_0}\bigl(\theta-\eta_{\mathrm{in}}\nabla_\theta \mathcal{L}_{\mathcal{D}_2}(\theta)\bigr).
\end{aligned}
\end{equation}

To expose the inductive bias induced by this objective, we apply a first-order Taylor approximation. Specifically, for a differentiable function $F(\theta)$, the first-order expansion around $\dot{\theta}$ is
\begin{equation}
\label{eq:taylor}
    F(\theta)
    \approx
    F(\dot{\theta})
    +
    \nabla_\theta F(\dot{\theta}) \cdot (\theta - \dot{\theta}).
\end{equation}

Applying this to each meta-learning regularization term yields
\[
\mathcal{L}_{\mathcal{D}_j}\!\left(\theta-\eta_{\text{in}}\nabla_\theta \mathcal{L}_{\mathcal{D}_i}(\theta)\right)
\approx
\mathcal{L}_{\mathcal{D}_j}(\theta)
-\eta_{\text{in}}\nabla_\theta \mathcal{L}_{\mathcal{D}_j}(\theta)\cdot \nabla_\theta \mathcal{L}_{\mathcal{D}_i}(\theta).
\]
Substituting this approximation into Equation~\ref{eqn:lmaml} gives
\begin{equation}
\label{eqn:mamlexpanded}
L_{\text{MAML}}(\theta)
\approx
2\Bigl(\mathcal{L}_{\mathcal{D}_0}(\theta)+\mathcal{L}_{\mathcal{D}_1}(\theta)+\mathcal{L}_{\mathcal{D}_2}(\theta)\Bigr)
+R_{01}(\theta)+R_{12}(\theta)+R_{20}(\theta),
\end{equation}

where
\begin{equation}
\label{eqn:alignment}
R_{ij}(\theta):=
-\eta_{\text{in}}\nabla_\theta \mathcal{L}_{D_i}(\theta)\cdot \nabla_\theta \mathcal{L}_{D_j}(\theta).
\end{equation}

The first component in Equation~\ref{eqn:mamlexpanded} corresponds to the standard supervised loss over $D_0$, $D_1$, and $D_2$. The remaining terms, $R_{01}(\theta)$, $R_{12}(\theta)$, and $R_{20}(\theta)$, are pairwise \textit{meta-learned gradient-alignment} terms. Specifically, $R_{ij}$ is the negative inner product between the gradient induced by the meta-train dataset $D_i$ and that induced by the corresponding meta-test dataset $D_j$. This dot product captures directional similarity, since for any two vectors $a$ and $b$, we have $a\cdot b=\|a\|_2\|b\|_2\cos(\delta)$, where $\delta$ denotes the angle between them. Consequently, minimizing the meta-learning objective implicitly \emph{encourages gradient alignment} across datasets\footnote{We omit the reverse task permutations $(D_1, D_0)$, $(D_2, D_1)$, and $(D_0, D_2)$ because, by symmetry of the gradient dot product, $R_{ij}=R_{ji}$, and therefore no qualitatively new terms are introduced for the analysis.}~\citep{li2017learninggeneralizemetalearningdomain}. Crucially, these alignment terms remain \emph{purely pairwise and independent} and do not encode how the inductive bias learned from an earlier pair should affect later query-induced objectives, and it is precisely the limitation that CAML addresses in Section~\ref{sec:caml_theory}.

\subsubsection{CAML and Cumulative Inductive Bias}
\label{sec:caml_theory}
We now derive the corresponding CAML objective under the same simplified two-round active learning setting. Unlike standard meta-learning, the round-2 adaptation in CAML is taken with respect to the round-1 objective $L_1$, so the second meta-learning term is explicitly conditioned on the inductive bias learned so far. Applying the CAML objective from Equation~\ref{eqn:CAML} gives
\begin{equation}
\begin{gathered}
L_0(\theta) = \mathcal{L}_{\mathcal{D}_0}(\theta), \\
L_1(\theta) = \mathcal{L}_{\mathcal{D}_0}(\theta) + \mathcal{L}_{\mathcal{D}_1}\!\left(\theta - \eta_{\text{in}} \nabla_\theta L_0(\theta)\right), \\
L_2(\theta) = \mathcal{L}_{\mathcal{D}_0}(\theta) + \mathcal{L}_{\mathcal{D}_1}\!\left(\theta - \eta_{\text{in}} \nabla_\theta L_0(\theta)\right) + \mathcal{L}_{\mathcal{D}_2}\!\left(\theta - \eta_{\text{in}} \nabla_\theta L_1(\theta)\right).
\end{gathered}
\end{equation}

Substituting the first-order Taylor approximations of the meta-learning terms into $L_2(\theta)$ yields

\begin{equation}
\label{eqn:l2_ultimate}
\begin{aligned}
L_2(\theta) \approx\;& \mathcal{L}_{\mathcal{D}_0}(\theta) + \mathcal{L}_{\mathcal{D}_1}(\theta) + \mathcal{L}_{\mathcal{D}_2}(\theta)
+ R_{01}(\theta) + R_{12}(\theta) + R_{20}(\theta) \\
& -\eta_{\text{in}} \bigl(\nabla_\theta R_{01}(\theta) \cdot \nabla_\theta \mathcal{L}_{\mathcal{D}_2}(\theta) + \nabla_\theta R_{01}(\theta) \cdot \nabla_\theta R_{12}(\theta) + \nabla_\theta R_{01}(\theta) \cdot \nabla_\theta R_{20}(\theta)\bigr).
\end{aligned}
\end{equation}

We defer the full derivation of Equation~\ref{eqn:l2_ultimate} to Appendix~B. Equation~\ref{eqn:l2_ultimate} is directly comparable to the standard MAML approximation in Equation~\ref{eqn:mamlexpanded}: both objectives contain the same supervised losses over $D_0$, $D_1$, and $D_2$, together with the same pairwise meta-learned gradient alignment terms $R_{01}(\theta)$, $R_{12}(\theta)$, and $R_{20}(\theta)$. The key difference is that CAML introduces additional \textit{inductive-bias interaction} terms that are absent from the independent-task formulation in standard meta-learning.

To interpret these terms, note that after round 1, the learned inductive bias is encoded by the alignment term $R_{01}(\theta)$. The second query batch introduces new objective components: the supervised loss $L_{D_2}(\theta)$ and the new pairwise alignment terms $R_{12}(\theta)$ and $R_{20}(\theta)$. The additional \textit{inductive-bias interaction} in Equation~\ref{eqn:l2_ultimate} therefore couples the previously learned inductive bias $R_{01}(\theta)$ with these newly introduced round-2 objective components. Because each interaction term is the negative dot product between the corresponding gradients, minimizing the CAML objective \emph{encourages alignment between the previously learned inductive bias and the new objectives}. This distinction is especially important in active learning. At round $i$, the query batch $D_i$ is selected by an acquisition function applied to the model trained under the previous objective $L_{i-1}$. Since $L_{i-1}$ already reflects the cumulative meta-learned regularization from earlier rounds, the resulting query batch is informative relative to the inductive bias learned so far, and how it should be updated. This dependence is precisely captured by the inductive-bias interaction terms, which promote the incremental updates needed to refine the previously accumulated inductive bias toward the current query batch.

%% file: Sections/F_Experiments.tex
\section{Experiments}
\label{section:experiments}
\subsection{Experimental Setup}
\label{section:setup}
We evaluate CAML on four spurious-correlation benchmarks: MNIST-CIFAR Dominoes~\citep{NEURIPS2020_6cfe0e61}, Waterbirds~\citep{DBLP:journals/corr/abs-1911-08731}, SpuCo~\citep{joshi2023mitigating}, and CivilComments~\citep{borkan2019nuancedmetricsmeasuringunintended, koh2021wildsbenchmarkinthewilddistribution}. Consistent with the broader spurious correlation literature~\citep{DBLP:journals/corr/abs-1911-08731, yang2023change, ye2024spurious}, we evaluate performance using \textit{minority-group test accuracy}, defined as accuracy on examples for which the spurious correlation does not hold. All results are reported as mean and standard deviation over three random seeds.
We use a pretrained ResNet-18 and a BERT model for vision and NLP datasets, respectively. For active learning, we consider four representative and widely used acquisition functions: Random (RAND), Confidence (CONF)~\citep{6889457}, CoreSet (CORE)~\citep{sener2018active}, and Batch Active learning by Diverse Gradient Embeddings (BADGE)~\citep{ash2020deep}. We initialize with $N_{L}=4500$ labeled examples and query in batches of $N_{B}=100$ until reaching a total labeled budget of $N_{max}=5000$. We compare CAML against the standard active learning pipeline, which simply adds queried samples back into the labeled set for subsequent training. We do not compare against methods discussed in Section~\ref{section:prior_spurious} that require privileged information, such as group or spurious-attribute annotations, since our focus is the standard active learning setting, which aims to automatically identify and mitigate spurious correlations through strategic data acquisition alone. The dataset details can be found in Appendix C; the experimental setup and its justification in Appendix D; the computational cost in Appendix E.

\subsection{Results}

\begin{table*}[tb]
    \caption{Final minority-group test accuracy (\%) across spurious-correlation benchmarks and acquisition functions, CAML against the standard active-learning pipeline.}
    \centering
    \resizebox{\textwidth}{!}{
    \begin{tabular}{c c c c c c c c c}
        \toprule
        \textbf{Acquisition} & 
        \multicolumn{2}{c}{\(\mathbf{Random}\)} & 
        \multicolumn{2}{c}{\(\mathbf{Confidence}\)} & 
        \multicolumn{2}{c}{\(\mathbf{CoreSet}\)} & 
        \multicolumn{2}{c}{\(\mathbf{BADGE}\)} \\

        \cmidrule(lr){2-3} \cmidrule(lr){4-5} \cmidrule(lr){6-7} \cmidrule(lr){8-9}
        \diagbox{Dataset}{Method}
 & Shuffle & CAML & 
          Shuffle & CAML & 
          Shuffle & CAML & 
          Shuffle & CAML \\

        \toprule
        \(\mathbf{Dominoes}\) & 
        \(25.4 \pm 0.8\) & \(\mathbf{29.1 \pm 1.4}\) & 
        \(40.6 \pm 0.7\) & \(\mathbf{53.3 \pm 1.3}\) & 
        \(41.0 \pm 2.2\) & \(\underline{\mathbf{68.8 \pm 1.6}}\) & 
        \(40.5 \pm 1.3\) & \(\mathbf{54.9 \pm 1.5}\) \\
        \(\mathbf{Waterbirds}\) & 
        \(43.8 \pm 0.9\) & \(\mathbf{57.8 \pm 1.2}\) & 
        \(49.6 \pm 1.0\) & \(\mathbf{65.5 \pm 1.9}\) & 
        \(46.9 \pm 2.4\) & \(\mathbf{64.3 \pm 1.8}\) & 
        \(39.1 \pm 1.5\) & \(\underline{\mathbf{69.0 \pm 1.7}}\) \\
        \(\mathbf{SpuCo}\) & 
        \(33.4 \pm 2.6\) & \(\mathbf{35.3 \pm 1.6}\) & 
        \(30.5\pm1.3\) & \(\mathbf{38.8 \pm 2.9}\) & 
        \(31.8 \pm 2.2\) & \(\mathbf{38.3 \pm 2.1}\) & 
        \(33.5 \pm 3.1\) & \(\underline{\mathbf{47.8 \pm 2.8}}\) \\

        \(\mathbf{CivilComments}\) & 
        \(40.1\pm1.5\) & \(\mathbf{49.2\pm2.1}\) & 
        \(40.3\pm1.5\) & \(\mathbf{58.0\pm1.2}\) & 
        \(39.1\pm1.1\) & \(\underline{\mathbf{63.1\pm1.3}}\) & 
        \(40.3\pm0.9\) & \(\mathbf{62.7\pm2.2}\) \\
        \bottomrule
    \end{tabular}
    }
    \label{Table:main_results}
\end{table*}

\begin{figure}[tb]
    \centering
    \begin{subfigure}[t]{0.19\linewidth}
        \centering
        \includegraphics[width=\linewidth]{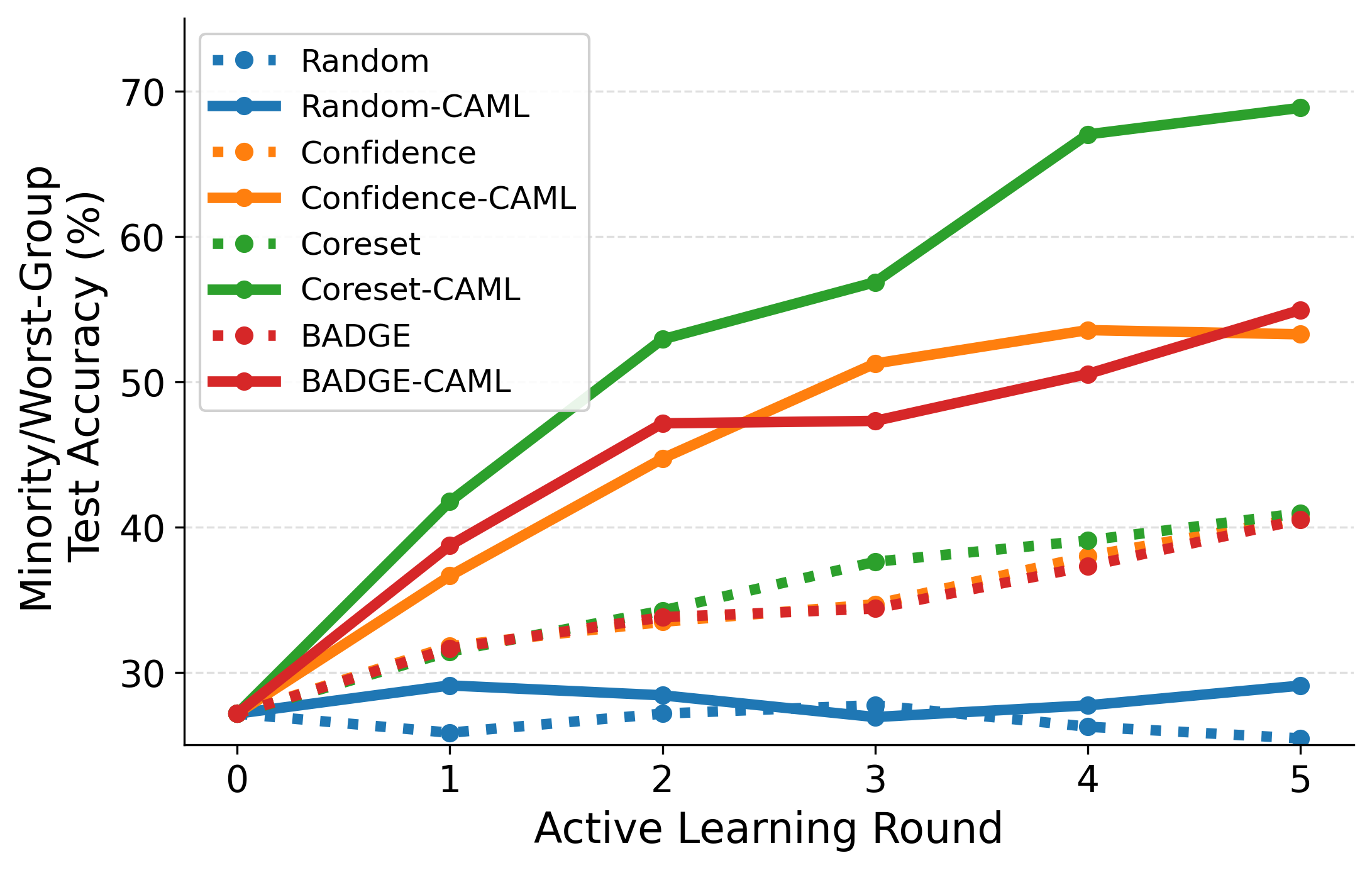}
        \caption{Dominoes}
        \label{fig:dominoes}
    \end{subfigure}\hfill
    \begin{subfigure}[t]{0.19\linewidth}
        \centering
        \includegraphics[width=\linewidth]{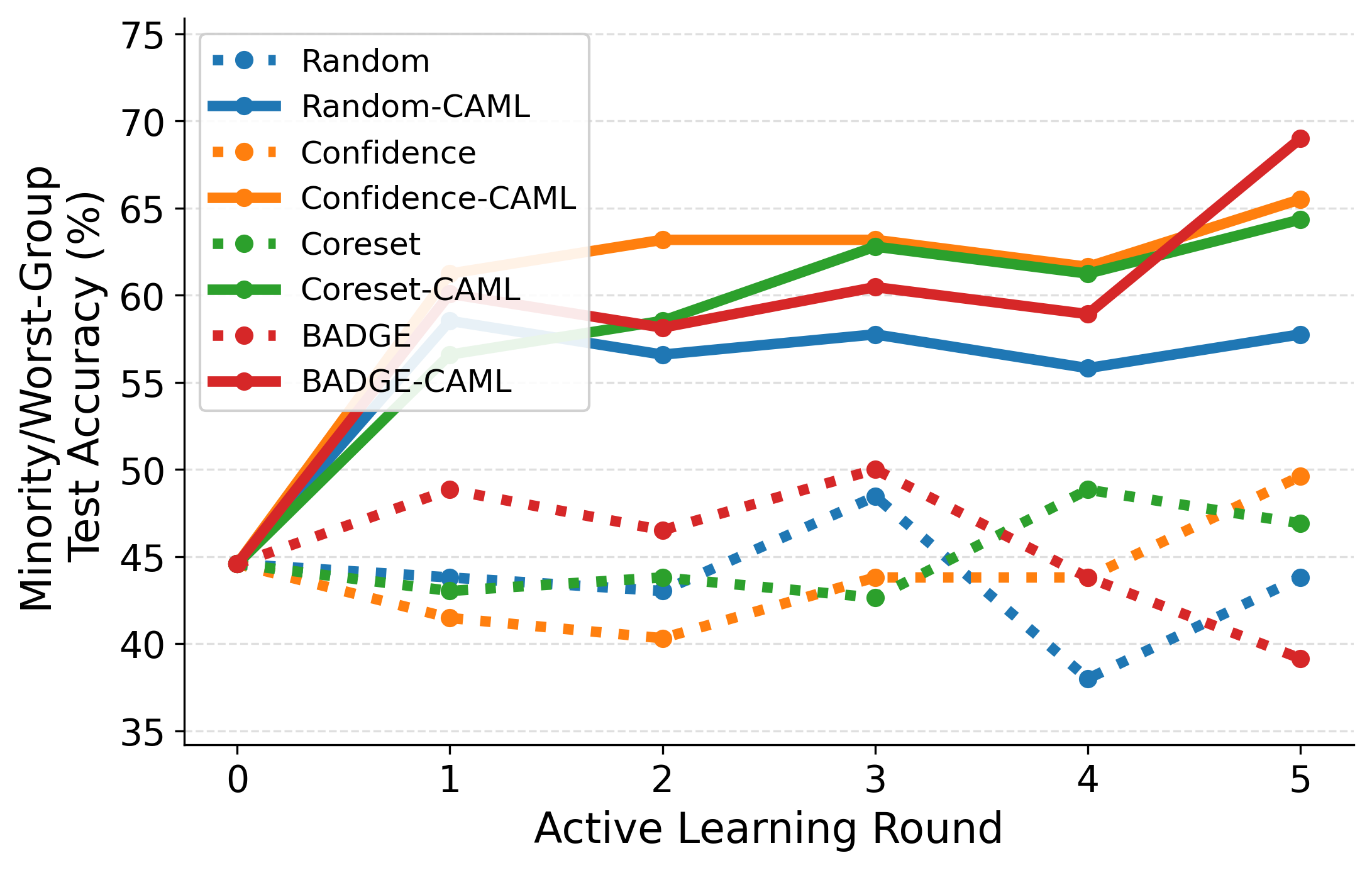}
        \caption{Waterbirds}
        \label{fig:waterbirds}
    \end{subfigure}\hfill
    \begin{subfigure}[t]{0.19\linewidth}
        \centering
        \includegraphics[width=\linewidth]{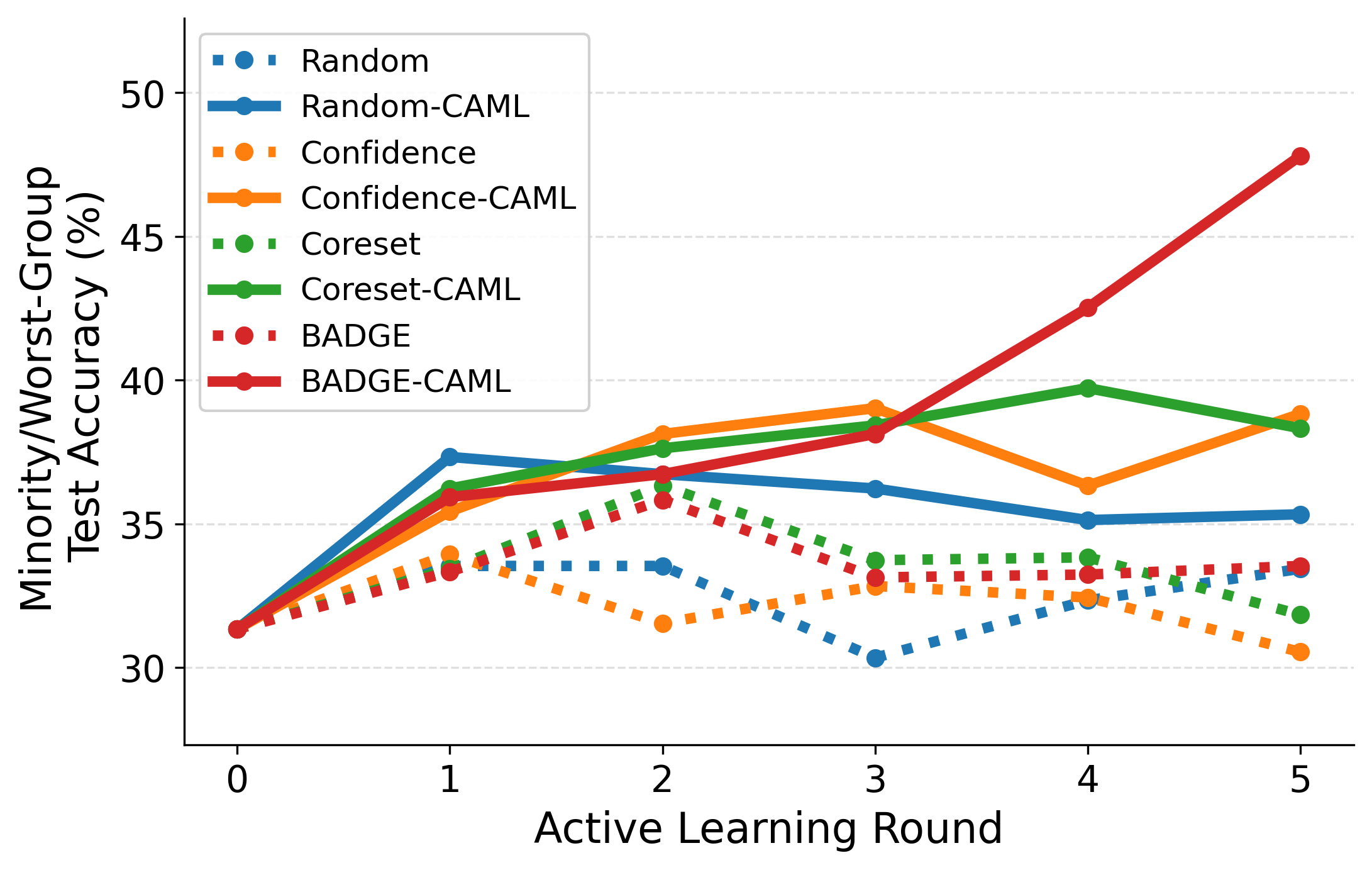}
        \caption{SpuCo}
        \label{fig:spuco}
    \end{subfigure}\hfill
    \begin{subfigure}[t]{0.19\linewidth}
        \centering
        \includegraphics[width=\linewidth]{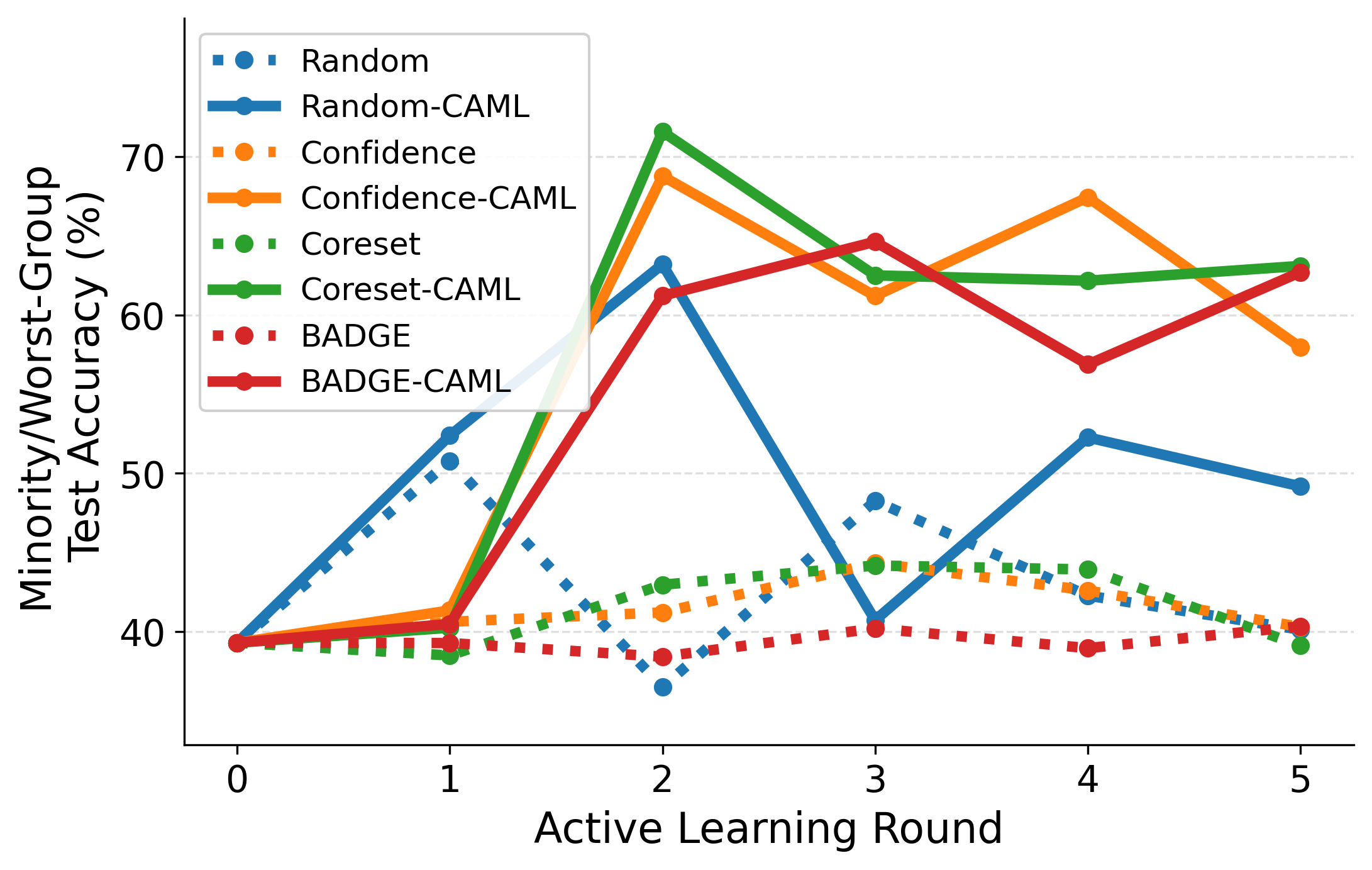}
        \caption{CivilComments}
        \label{fig:civilcomments}
    \end{subfigure}\hfill
    \begin{subfigure}[t]{0.19\linewidth}
        \centering
        \includegraphics[width=\linewidth]{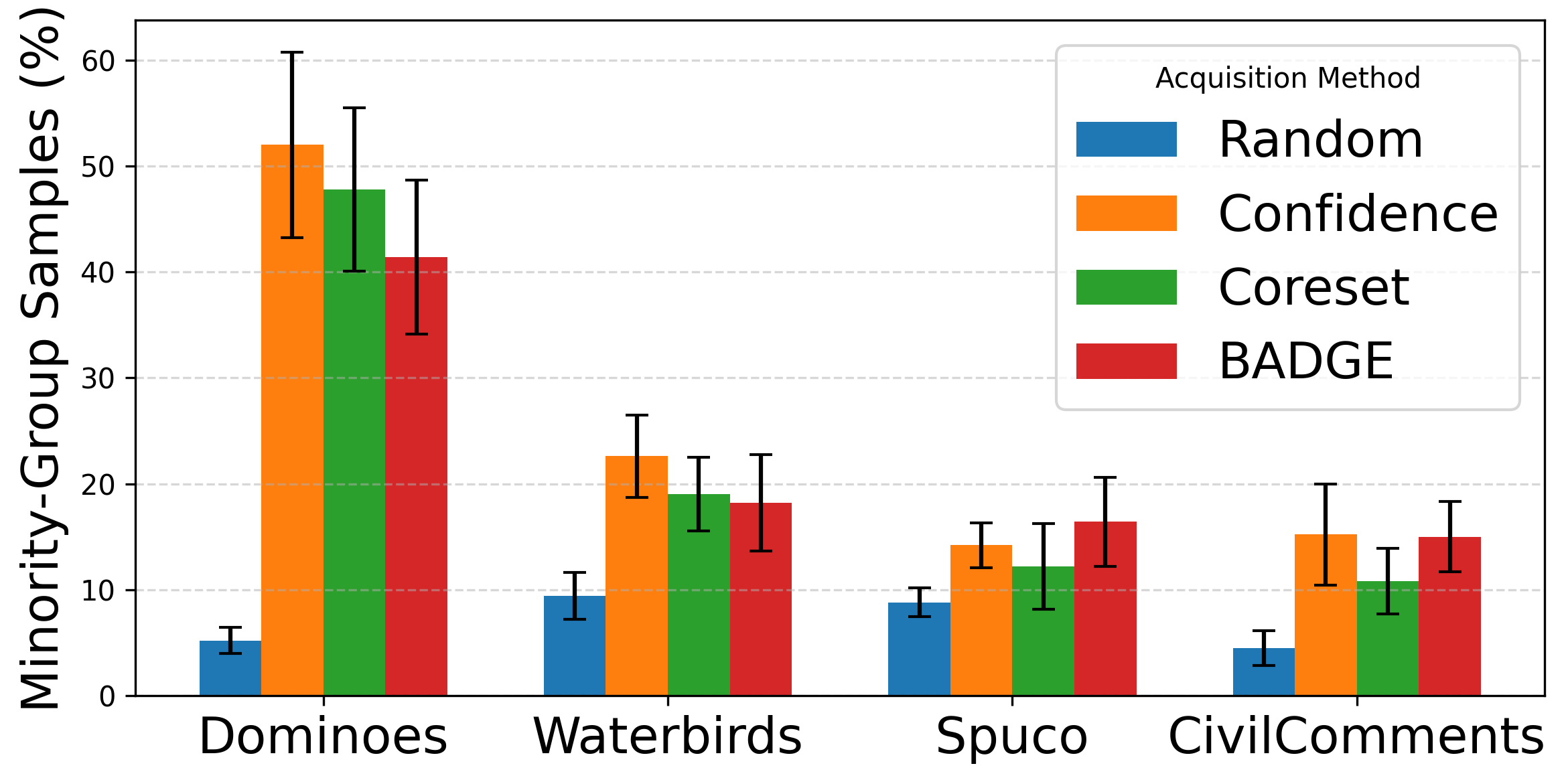}
        \caption{Minority Query}
        \label{fig:minority_query}
    \end{subfigure}\hfill
    \caption{Minority-group test accuracy (\%) across active-learning rounds across spurious-correlation benchmarks and acquisition functions, comparing CAML against the standard active-learning pipeline. Panel (e) shows the average proportion of minority groups in each queried batch.}
    \label{fig:acquisition_results}
\end{figure}

\textbf{Impact of Acquisition Functions.} We observe that informed acquisition functions generally outperform random sampling through the selection of the more informative minority-group samples, which are necessary to disprove spurious correlations in the training data. Figure~\ref{fig:acquisition_results} illustrates the larger performance gains of informed acquisition functions over random sampling across active-learning rounds, with the final Dominoes accuracies in Table~\ref{Table:main_results} showing that Confidence, CoreSet, and BADGE achieve $40.6\%$, $41.0\%$, and $40.5\%$, respectively, compared to $25.4\%$ for random sampling. This trend is partly explained by Figure~\ref{fig:minority_query}, where the Confidence, CoreSet, and BADGE acquisition strategies query substantially higher proportions of minority-group samples ($35.80\%$, $54.40\%$, and $30.00\%$, respectively), compared to only $3.60\%$ for random sampling. While the minority-group sampling rate is not by itself a perfect predictor of final performance, it serves as a valuable proxy for acquisition quality, since these samples are essential for learning core rather than spurious features. These findings are consistent with prior observations by~\citet{tamkin2022active}.

\textbf{CAML's Performance.} We observe that CAML consistently improves over the standard Shuffle baseline across informed acquisition functions and datasets, achieving gains of up to $27.8\%$, $29.9\%$, $14.3\%$, and $24.0
\%$ respectively (Table~\ref{Table:main_results}). Figure~\ref{fig:acquisition_results} visualizes this effect, where CAML generally remains above the Shuffle baseline throughout active-learning rounds. These improvements support the claim that CAML can more effectively exploit informative query data to shape the model's inductive bias. On spurious-correlation datasets, this guides the model toward core features while reducing reliance on spurious correlations, thereby improving minority-group performance.

\subsection{Ablations}
We follow the experimental setup described in Section~\ref{section:setup} and perform all ablation studies on a single representative setting: BADGE on the MNIST-CIFAR Dominoes dataset. We ablate the effects of three key parameters: weight decay, query batch informativeness, and active learning round.

\begin{figure}[tb]
    \centering
    \begin{subfigure}[t]{0.25\linewidth}
        \centering
        \includegraphics[width=\linewidth]{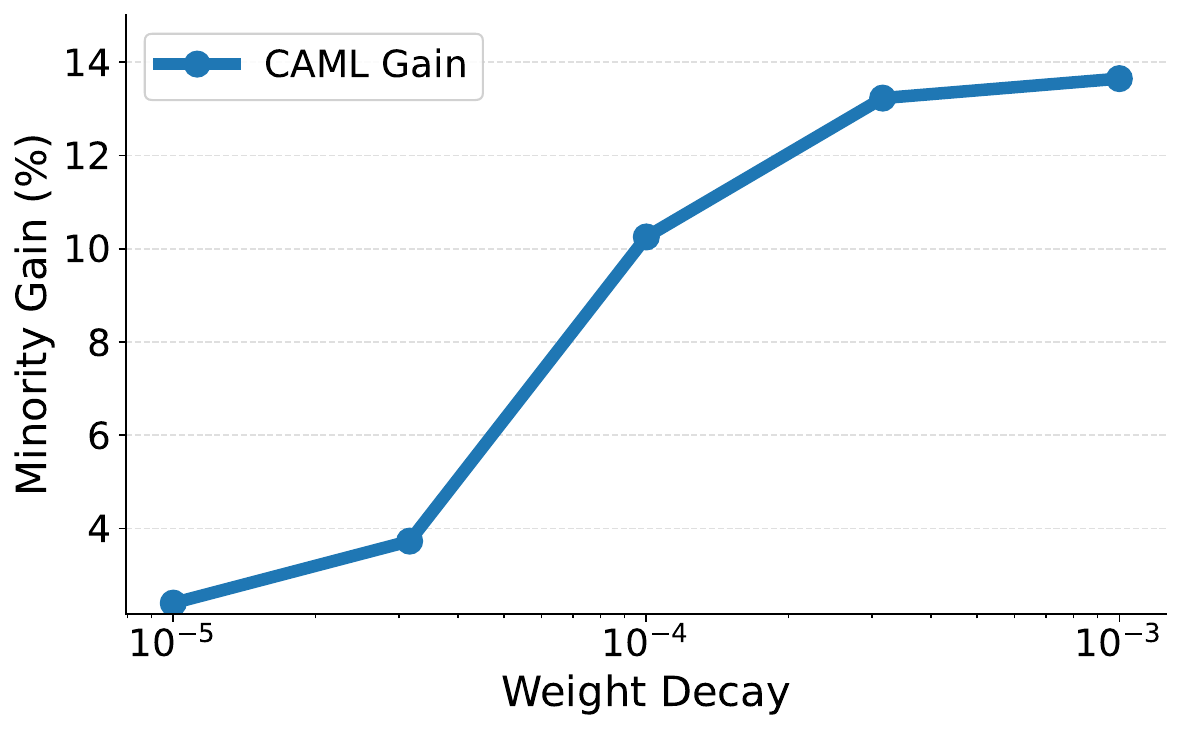}
    \end{subfigure}
    \begin{subfigure}[t]{0.25\linewidth}
        \centering
        \includegraphics[width=\linewidth]{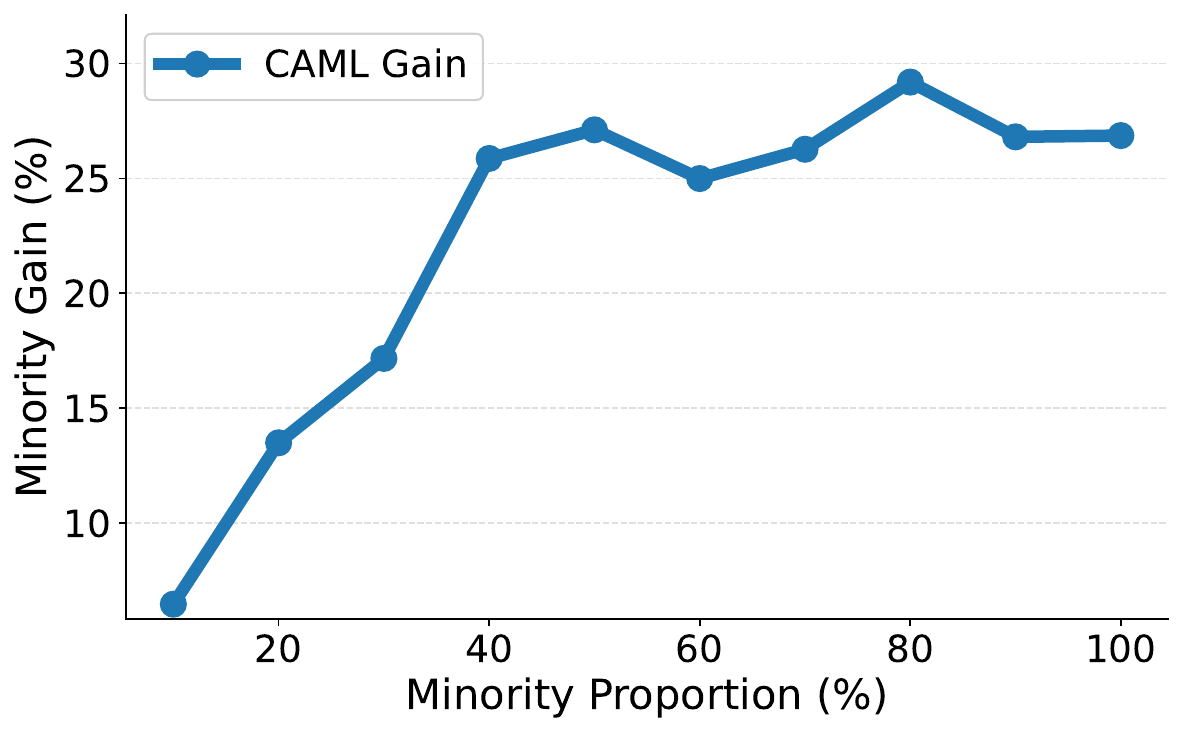}
    \end{subfigure}
    \begin{subfigure}[t]{0.25\linewidth}
        \centering
        \includegraphics[width=\linewidth]{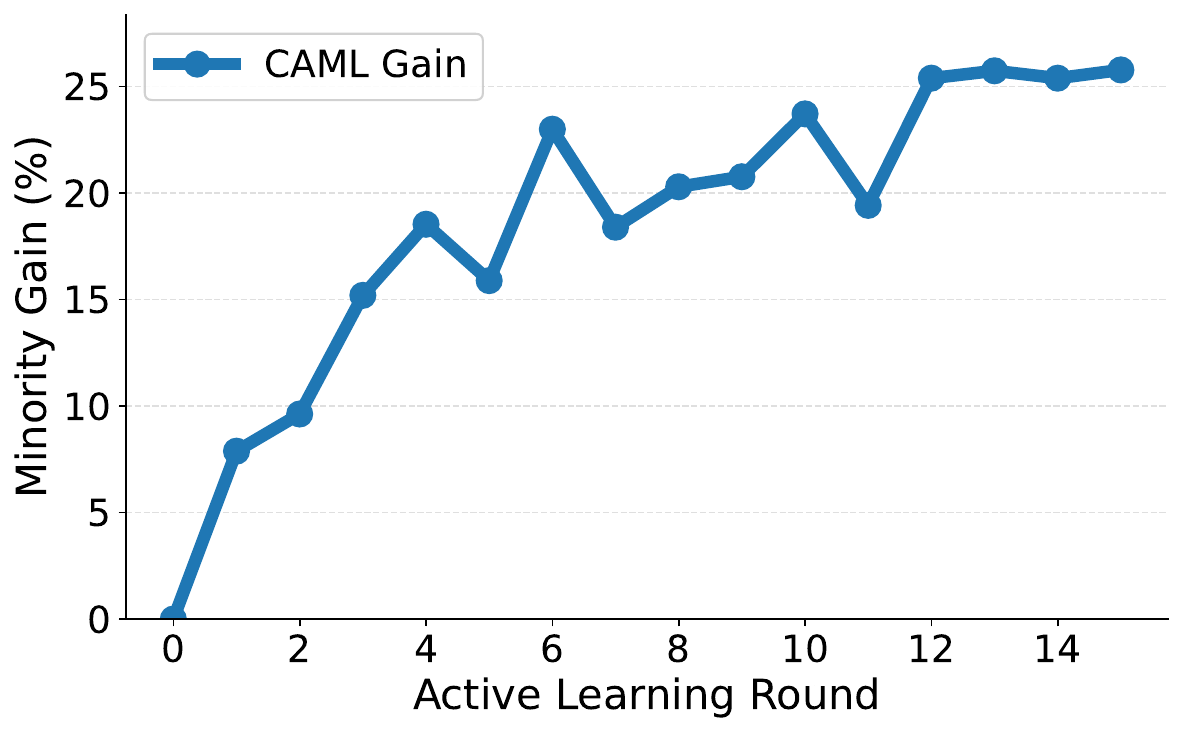}
    \end{subfigure}
    \caption{Ablation studies characterizing CAML's minority-group accuracy gains over the Shuffle baseline on Dominoes with BADGE acquisition. CAML achieves larger gains under \textbf{(a)} stronger regularization, \textbf{(b)} higher minority-query proportions, and \textbf{(c)} more active learning rounds.}
    \label{fig:ablation}
\end{figure}

\paragraph{Weight Decay.} Figure~\ref{fig:ablation}a shows that CAML's advantage becomes more pronounced as weight decay increases. This is consistent with prior work showing that strong regularization is necessary for minority-group generalization in the overparameterized regime to constrain model capacity, reducing memorization of minority-group examples~\citep{DBLP:journals/corr/abs-1911-08731, pmlr-v119-sagawa20a}.

\paragraph{Query Batch Informativeness.}
Figure~\ref{fig:ablation}b shows that CAML’s advantage increases as the proportion of minority-group examples in each query increases. This trend supports CAML's core principle: amplifying informative queried batches, which in spurious-correlation settings correspond to minority-group examples that violate the spurious correlation.

\paragraph{Active Learning Round.} Figure~\ref{fig:ablation}c shows that CAML's advantage widens across active-learning rounds, suggesting that its cumulative inductive bias is being progressively refined.

%% file: Sections/G_Limitations.tex
\section{Limitations and Future Work}
\label{section:CAML_limitations}
First, CAML's gains depend on the acquisition function's ability to select informative queried examples, highlighting the synergistic relationship between acquisition design and query integration. Second, CAML incurs additional computational cost due to the cumulative meta-learning objective, although our first-order implementation avoids expensive second-order derivatives. Finally, although our main empirical evaluation focuses on standard spurious-correlation benchmarks, we include an additional non-spurious CIFAR100 experiment in Appendix~E; extending CAML more systematically to broader datasets and distribution-shift settings remains an important direction for future work.

%% file: Sections/H_Conclusion.tex
\section{Conclusion}
\label{section:conclusion}

We introduced Cumulative Active Meta-Learning (CAML), a framework for improving robustness to spurious correlations by using actively queried examples to refine the model's inductive bias.  Unlike standard active-learning pipelines that simply append queried samples to the labeled set, CAML treats each queried batch as a meta-test signal that shapes how the model adapts to the labeled data. Across sequentially dependent active learning rounds, the inductive bias is cumulated and refined. Our theoretical analysis shows that this cumulative formulation captures dependencies absent from conventional independent-task meta-learning, and our experiments demonstrate consistent gains in minority-group accuracy across spurious-correlation benchmarks and acquisition strategies.

%% file: Appendix/A_Method.tex
\section{Efficient Implementation of CAML}
\label{app:caml_algorithm}

\paragraph{Objective Function:} Equation~\ref{eqn:CAML} defines the CAML objective at active learning
round $n$ as
\[
L_n(\theta)
=
\mathcal{L}_{\mathcal{D}_0}(\theta)
+
\sum_{i=1}^{n}
\mathcal{L}_{\mathcal{D}_i}
\left(
\theta - \eta_{\mathrm{in}} \nabla_\theta L_{i-1}(\theta)
\right),
\]
where $\mathcal{D}_0$ is the initial labeled set and $\mathcal{D}_i$ is the query batch acquired at
round $i$.

\paragraph{Base Case:} Let $G_i$ denote the first-order estimate of the cumulative gradient
$\nabla_\theta L_i(\theta)$. We initialize
\[
G_0 = \nabla_\theta L_{0}(\theta) = \nabla_\theta \mathcal{L}_{\mathcal{D}_0}(\theta).
\]
\paragraph{Forward Recursion:} For each queried batch $D_i$, CAML simulates the one-step adaptation on the previous cumulative objective $L_{i-1}$ using the cumulative gradient $G_{i-1}$
\[
\tilde{\theta}
=
\theta - \eta_{\mathrm{in}} G_{i-1},
\]
The queried batch $D_i$ is then evaluated at $\tilde{\theta}$. We compute the gradient of this query loss directly with respect to $\tilde{\theta}$ to obtain the first-order approximation of the $i$-th CAML gradient term that is cumulated to obtain $G_i$
:
\[
G_i
=
G_{i-1}
+
\nabla_{\tilde{\theta}}
\mathcal{L}_{D_i}(\tilde{\theta}).
\]
This yields a forward recursion over active-learning rounds. After processing all queried batches, we obtain $G_n$, which represents the first-order estimate of the cumulative gradient $\nabla_\theta L_n(\theta)$ and is used to update the model.

\paragraph{Computational Complexity:} CAML avoids second-order derivatives and requires only $n+1$ forward/backward passes per optimization step: one for the initial labeled set $D_0$ and one for each queried batch $D_1,\ldots,D_n$.

\begin{algorithm}[h]
\caption{Efficient First-Order Implementation of CAML}
\label{alg:caml}
\begin{algorithmic}[1]
\REQUIRE Initial labeled set $D_0$; queried batches $\{D_i\}_{i=1}^{n}$; model $f_\theta$; loss function $\ell$; optimizer $\mathrm{opt}$; inner step size $\eta_{\mathrm{in}}$
\FOR{epoch $=1,\ldots,T$}
    \FOR{each optimization step}
        \STATE Sample mini-batches $B_0 \sim D_0$ and $B_i \sim D_i$ for $i=1,\ldots,n$
        \STATE $\mathrm{opt.zero\_grad}()$
        \STATE $G_0 \leftarrow \nabla_\theta \mathcal{L}_{B_0}(\theta)$
        \FOR{$i=1,\ldots,n$}
            \STATE $\tilde{\theta} \leftarrow \theta - \eta_{\mathrm{in}} G_{i-1}$
            \COMMENT{simulated update using $G_{i-1}$}
            \STATE $G_i \leftarrow G_{i-1} +  \nabla_{\tilde{\theta}} \mathcal{L}_{B_i}(\tilde{\theta})$ \COMMENT{forward recursion to compute $G_i$}
        \ENDFOR
        \STATE Set $\theta.\mathrm{grad} \leftarrow G_n$
        \STATE $\mathrm{opt.step}()$
    \ENDFOR
\ENDFOR
\STATE \RETURN $f_\theta$
\end{algorithmic}
\end{algorithm}

%% file: Appendix/B_TheoreticalProof.tex
\section{Theoretical Proof}

This appendix derives the closed-form approximation in Equation~\ref{eqn:l2_ultimate} from the CAML objective in Equation~\ref{eqn:CAML}, under the simplified two-round active learning setting described in Section~\ref{section:theoretical_analysis}. We consider an initial labeled dataset $D_0$ and two queried batches $D_1$ and $D_2$, obtained in the first and second rounds, respectively. The derivation applies the first-order Taylor approximation in Equation~\ref{eq:taylor} to the round-2 CAML objective. For completeness, we restate the relevant equations below.

We begin from the CAML objective in Equation~\ref{eqn:CAML}:
\begin{equation*}
    L_n(\theta)
    =
    \mathcal{L}_{D_0}(\theta)
    +
    \sum_{i=1}^{n}
    \mathcal{L}_{D_i}\bigl(\theta - \eta_{\mathrm{in}}\,\nabla_{\theta} L_{i-1}(\theta)\bigr).
\end{equation*}

We then apply the first-order Taylor approximation from Equation~\ref{eq:taylor}:
\begin{equation*}
    F(\theta)
    \approx
    F(\dot{\theta})
    +
    \nabla_\theta F(\dot{\theta}) \cdot (\theta - \dot{\theta}).
\end{equation*}

Finally, we use the gradient alignment regularization term from Equation~\ref{eqn:alignment} as shorthand:
\begin{equation*}
    R_{ij}(\theta)
    :=
    - \eta_{\mathrm{in}} \,
    \nabla_\theta \mathcal{L}_{D_i}(\theta)
    \cdot
    \nabla_\theta \mathcal{L}_{D_j}(\theta).
\end{equation*}

\paragraph{Round 0:}
At initialization, the cumulative CAML objective reduces to the supervised loss on the initial labeled dataset:
\begin{equation}
    \label{eqn:l0}
    L_0(\theta) = \mathcal{L}_{D_0}(\theta).
\end{equation}
This serves as the base case for the recursive construction of the CAML objective in later rounds.

\paragraph{Round 1:}
At round 1, the CAML objective consists of the supervised loss on the initial labeled dataset and one meta-learning regularization term on the first queried batch:
\begin{equation*}
    L_1(\theta)
    =
    \mathcal{L}_{D_0}(\theta)
    +
    \mathcal{L}_{D_1}\bigl(\theta - \eta_{\mathrm{in}} \nabla_\theta L_0(\theta)\bigr).
\end{equation*}

Substituting the base case from Equation~\ref{eqn:l0} gives
\begin{equation*}
    L_1(\theta)
    =
    \mathcal{L}_{D_0}(\theta)
    +
    \mathcal{L}_{D_1}\bigl(\theta - \eta_{\mathrm{in}} \nabla_\theta \mathcal{L}_{D_0}(\theta)\bigr).
\end{equation*}

To isolate the derivation of the new meta-learning regularization term at round 1, define
\begin{equation*}
    F_1(\theta)
    =
    \mathcal{L}_{D_1}\bigl(\theta - \eta_{\mathrm{in}} \nabla_\theta \mathcal{L}_{D_0}(\theta)\bigr).
\end{equation*}

Applying the first-order Taylor approximation around $\dot{\theta} = \theta$ gives
\begin{equation*}
    F_1(\theta)
    \approx
    \mathcal{L}_{D_1}(\theta)
    -
    \eta_{\mathrm{in}}
    \nabla_\theta \mathcal{L}_{D_0}(\theta)
    \cdot
    \nabla_\theta \mathcal{L}_{D_1}(\theta).
\end{equation*}

Using the definition of the pairwise alignment term in Equation~\ref{eqn:alignment}, this becomes
\begin{equation}
    \label{eqn:l1}
    F_1(\theta)
    \approx
    \mathcal{L}_{D_1}(\theta) + R_{01}(\theta).
\end{equation}

Therefore, the round-1 objective simplifies to
\begin{equation}
    \label{eqn:l1final}
    L_1(\theta)
    \approx
    \mathcal{L}_{D_0}(\theta)
    +
    \mathcal{L}_{D_1}(\theta)
    +
    R_{01}(\theta).
\end{equation}

\paragraph{Round 2:} At round 2, the CAML objective consists of the round-1 objective together with a new meta-learning regularization term on the second queried batch:

\begin{equation*}
    L_2(\theta)
    =
    \mathcal{L}_{D_0}(\theta)
    +
    \mathcal{L}_{D_1}\bigl(\theta - \eta_{\mathrm{in}} \nabla_\theta L_0(\theta)\bigr)
    +
    \mathcal{L}_{D_2}\bigl(\theta - \eta_{\mathrm{in}} \nabla_\theta L_1(\theta)\bigr).
\end{equation*}
Substituting the round-0 and round-1 expansions from Equations~\ref{eqn:l0}, \ref{eqn:l1}, and \ref{eqn:l1final} gives

\begin{equation}
    \label{eqn:l2}
    L_2(\theta)
    \approx
    \mathcal{L}_{D_0}(\theta)
    +
    \mathcal{L}_{D_1}(\theta)
    +
    R_{01}(\theta)
    +
    \mathcal{L}_{D_2}\Bigl(
        \theta
        -
        \eta_{\mathrm{in}}
        \nabla_\theta
        \bigl(
            \mathcal{L}_{D_0}(\theta)
            +
            \mathcal{L}_{D_1}(\theta)
            +
            R_{01}(\theta)
        \bigr)
    \Bigr).
\end{equation}

To isolate the derivation of the new meta-learning regularization term at round 2, define
\begin{equation*}
    \label{eqn:f2}
    F_2(\theta)
    =
    \mathcal{L}_{D_2}\Bigl(
        \theta
        -
        \eta_{\mathrm{in}}
        \nabla_\theta
        \bigl(
            \mathcal{L}_{D_0}(\theta)
            +
            \mathcal{L}_{D_1}(\theta)
            +
            R_{01}(\theta)
        \bigr)
    \Bigr).
\end{equation*}

Its inner update can be decomposed into the contribution from the previous supervised-loss terms and the contribution from the previously learned regularization term $R_{01}(\theta)$:

\begin{equation*}
    F_2(\theta)
    =
    \mathcal{L}_{D_2}\Bigl(
        \theta
        -
        \eta_{\mathrm{in}}
        \nabla_\theta
        \bigl(
            \mathcal{L}_{D_0}(\theta)
            +
            \mathcal{L}_{D_1}(\theta)
        \bigr)
        -
        \eta_{\mathrm{in}}
        \nabla_\theta R_{01}(\theta)
    \Bigr).
\end{equation*}

Applying the first-order Taylor approximation around $
    \dot{\theta}
    =
    \theta
    -
    \eta_{\mathrm{in}}
    \nabla_{\theta}
    \bigl(
        \mathcal{L}_{D_0}(\theta)
        +
        \mathcal{L}_{D_1}(\theta)
    \bigr)
$
gives

\begin{equation}
    \label{eqn:l2_outer}
    \begin{aligned}
        F_2(\theta)
        &\approx
        \mathcal{L}_{D_2}\Bigl(
            \theta
            -
            \eta_{\mathrm{in}}
            \nabla_{\theta}
            \bigl(
                \mathcal{L}_{D_0}(\theta)
                +
                \mathcal{L}_{D_1}(\theta)
            \bigr)
        \Bigr) \\
        &\quad
        -
        \eta_{\mathrm{in}}
        \nabla_{\theta}
        \mathcal{L}_{D_2}\Bigl(
            \theta
            -
            \eta_{\mathrm{in}}
            \nabla_{\theta}
            \bigl(
                \mathcal{L}_{D_0}(\theta)
                +
                \mathcal{L}_{D_1}(\theta)
            \bigr)
        \Bigr)
        \cdot
        \nabla_{\theta}R_{01}(\theta).
    \end{aligned}
\end{equation}

We now isolate the meta-learning regularization term with only the supervised-loss inner update:

\begin{equation*}
    F'_2(\theta)
    =
    \mathcal{L}_{D_2}\Bigl(
        \theta
        -
        \eta_{\mathrm{in}}
        \nabla_{\theta}
        \bigl(
            \mathcal{L}_{D_0}(\theta)
            +
            \mathcal{L}_{D_1}(\theta)
        \bigr)
    \Bigr).
\end{equation*}

Applying the first-order Taylor approximation around $\dot{\theta} = \theta$ gives
\begin{equation*}
    \begin{aligned}
        F'_2(\theta)
        &\approx
        \mathcal{L}_{D_2}(\theta)
        -
        \eta_{\mathrm{in}}
        \nabla_{\theta}\mathcal{L}_{D_2}(\theta)
        \cdot
        \nabla_{\theta}
        \bigl(
            \mathcal{L}_{D_0}(\theta)
            +
            \mathcal{L}_{D_1}(\theta)
        \bigr) \\
        &=
        \mathcal{L}_{D_2}(\theta)
        -
        \eta_{\mathrm{in}}
        \nabla_{\theta}\mathcal{L}_{D_2}(\theta)
        \cdot
        \nabla_{\theta}\mathcal{L}_{D_0}(\theta)
        -
        \eta_{\mathrm{in}}
        \nabla_{\theta}\mathcal{L}_{D_2}(\theta)
        \cdot
        \nabla_{\theta}\mathcal{L}_{D_1}(\theta) 
    \end{aligned}
\end{equation*}
Using the definition of the pairwise alignment term in Equation~\ref{eqn:alignment}, this becomes
\begin{equation}
    \label{eqn:l2inner}
    F'_2(\theta)
        \approx \mathcal{L}_{D_2}(\theta)
        +
        R_{20}(\theta)
        +
        R_{12}(\theta).
\end{equation}

Substituting Equation~\ref{eqn:l2inner} into Equation~\ref{eqn:l2_outer} gives
\begin{equation}
    \label{eqn:l2_inner_final}
    \begin{aligned}
        F_2(\theta)
        &\approx
        \mathcal{L}_{D_2}(\theta)
        +
        R_{20}(\theta)
        +
        R_{12}(\theta) \\
        &\quad
        -
        \eta_{\mathrm{in}}
        \nabla_\theta
        \Bigl(
            \mathcal{L}_{D_2}(\theta)
            +
            R_{20}(\theta)
            +
            R_{12}(\theta)
        \Bigr)
        \cdot
        \nabla_{\theta}R_{01}(\theta) \\
        &=
        \mathcal{L}_{D_2}(\theta)
        +
        R_{20}(\theta)
        +
        R_{12}(\theta)
        - \\
        &\quad
        \eta_{\mathrm{in}}
        \nabla_\theta\mathcal{L}_{D_2}(\theta)
        \cdot
        \nabla_{\theta}R_{01}(\theta) 
        -
        \eta_{\mathrm{in}}
        \nabla_\theta R_{20}(\theta)
        \cdot
        \nabla_{\theta}R_{01}(\theta)
        -
        \eta_{\mathrm{in}}
        \nabla_\theta R_{12}(\theta)
        \cdot
        \nabla_{\theta}R_{01}(\theta).
    \end{aligned}
\end{equation}

Finally, substituting Equation~\ref{eqn:l2_inner_final} into Equation~\ref{eqn:l2} gives Equation~\ref{eqn:l2_ultimate}

\begin{equation*}
    \begin{aligned}
        L_2(\theta)
        &\approx
        \mathcal{L}_{D_0}(\theta)
        +
        \mathcal{L}_{D_1}(\theta)
        +
        \mathcal{L}_{D_2}(\theta)
        +
        R_{01}(\theta)
        +
        R_{12}(\theta)
        +
        R_{20}(\theta) \\
        &\quad
        -
        \eta_{\mathrm{in}}
        \nabla_\theta\mathcal{L}_{D_2}(\theta)
        \cdot
        \nabla_{\theta}R_{01}(\theta)
        -
        \eta_{\mathrm{in}}
        \nabla_\theta R_{20}(\theta)
        \cdot
        \nabla_{\theta}R_{01}(\theta)
        -
        \eta_{\mathrm{in}}
        \nabla_\theta R_{12}(\theta)
        \cdot
        \nabla_{\theta}R_{01}(\theta).
    \end{aligned}
\end{equation*}

%% file: Appendix/C_Datasets.tex
\section{Datasets}
Similar to \citep{DBLP:journals/corr/abs-2110-14503}, the breakdown of the training split group counts for the spurious correlation datasets is presented in Table \ref{Table:train_group_counts_simple}. Notably, the CivilComments dataset is modified to set the spurious correlation strength to $95\%$ and remove class imbalance.

\begin{table*}[ht]
\caption{Training split group counts for the spurious correlation datasets. The dataset is constructed so that each label is uniquely correlated with a single spurious attribute, which the model can exploit as a shortcut to minimize empirical risk. The counts for the majority groups are listed along the diagonal, while the counts for the minority groups are shown off-diagonal.}
\label{Table:train_group_counts_simple}
\centering
\begin{tabular}{ c c c c c c c} 
\hline

\textbf{Dataset} & \textbf{Label} & \multicolumn{5}{c}{\textbf{Train Group Counts}} \\

\hline
 & \(\downarrow\) \textit{y} \(\rightarrow\) \textit{a} & Digit 0 & Digit 1 &\(\cdots\) & Digit 8 &Digit 9 \\
\cmidrule(lr){3-7} 

\multirow{5}{*}{\textbf{MNIST-CIFAR Dominoes}}&Airplane & 4085 & 24 & \(\cdots\) & 24 & 24\\
&Automobile & 24 & 4800 & \(\cdots\) & 24 & 24\\

&\(\vdots\) & \(\vdots\) & \(\vdots\) & \(\ddots\) & \(\vdots\)  & \(\vdots\)\\
&Ship & 24 & 24 & \(\cdots\) & 4800 & 24 \\

&Truck & 24 & 24 & \(\cdots\) & 24 & 4800 
\\

\midrule
& \(\downarrow\) \textit{y} \(\rightarrow\) \textit{a} & Land & Water & \\
\cmidrule(lr){3-4}
\multirow{2}{*}{\textbf{Waterbirds}}& Landbirds & 4500 & 236  \\
&Waterbirds & 236 & 4500 \\ 

\midrule
& \(\downarrow\) \textit{y} \(\rightarrow\) \textit{a} & Land & Water &Indoors & Outdoors & \\
\cmidrule(lr){3-6}
\multirow{4}{*}{\textbf{SpuCo}}& Landbirds & 4500 & 236 &0 & 0 & \\
&Waterbirds & 236 & 4500 & 0 & 0 & \\ 
&Small Dogs & 0 & 0 & 4500 & 236 &  \\
&Big Dogs & 0 & 0 & 236 & 4500 &  \\
\midrule
& \(\downarrow\) \textit{y} \(\rightarrow\) \textit{a} & Identity & Other\\
\cmidrule(lr){3-4}
\multirow{2}{*}{\textbf{CivilComments}}& Toxic & 15000 & 800 \\
&Non-Toxic & 800 & 15000 \\ 

\hline
\end{tabular}
\end{table*}

%% file: Appendix/D_ExperimentalSetup.tex
\section{Experimental Setup}
\paragraph{Spurious Correlation Setup} We conduct experiments on the MNIST-CIFAR Dominoes \citep{NEURIPS2020_6cfe0e61}, Waterbirds~\citep{DBLP:journals/corr/abs-1911-08731}, SpuCo \citep{joshi2023mitigating}, and CivilComments \citep{borkan2019nuancedmetricsmeasuringunintended, koh2021wildsbenchmarkinthewilddistribution} datasets. Details on the dataset are provided in Appendix C. Consistent with the broader spurious correlation literature~\citep{DBLP:journals/corr/abs-1911-08731, yang2023change, ye2024spurious}, we evaluate performance using minority-group test accuracy, i.e., accuracy on examples for which the spurious correlation does not hold. By contrast, a high majority-group or overall accuracy can often be achieved by relying on the spurious feature.

\paragraph{\textbf{Training Setup}} Following a similar setup to \citet{DBLP:journals/corr/abs-2110-14503}, we use a pretrained ResNet-18 and BERT model for vision and NLP datasets respectively. All experiments are conducted using pre-trained models, following the findings of \citet{tamkin2022active}, who highlighted the importance of pre-training for active learning in settings with spurious correlations. We use default training hyper-parameters of betas = (0.9, 0.999), learning rate = $10^{-4}$, and batch size = $32$. Additionally, we use a high weight decay value of $10^{-2}$ since \citet{DBLP:journals/corr/abs-1911-08731, pmlr-v119-sagawa20a} found that large regularization is necessary for minority-group generalization in the overparameterized regime. No hyper-parameter tuning is performed as it is computationally expensive in the active learning setup \citep{ash2020deep}. Training is conducted for 100 epochs to ensure convergence. The best model, determined by the validation dataset performance, is saved for final evaluation.

\paragraph{\textbf{Active Learning Setup}} We select four representative and widely used acquisition functions: Random (RAND), Confidence (CONF), CoreSet (CORE), and Batch Active learning by Diverse Gradient Embeddings (BADGE). We start with $N_{L}=4500$ labeled data points and query in batches of $N_{B}=100$ data points until the dataset contains $N_{max}=5000$ data points. This setup enables us to evaluate how effectively the algorithms mitigate spurious correlations under limited data conditions, aligning with the goal of active learning of maximizing performance given a tight labeling budget. We present learning curves that plot test performance across the active learning rounds, along with the final test performance, both of which are commonly used metrics in active learning \citep{settles.tr09}. After each active learning round, models are retrained from fresh random initializations to prevent the warm-starting problem \citep{ash2020warmstarting}.

\paragraph{\textbf{Experiments}} We evaluate CAML's ability to more effectively utilize the actively queried samples compared to the standard method of shuffling them back into the dataset. Experiments are repeated with three different random seeds to ensure robustness. We exclude the methods discussed in Section \ref{section:prior_spurious}, as many of them rely on additional assumptions or external information—such as group annotations or spurious attribute labels—to resolve spurious correlations. This contrasts with the goal of active learning, which aims to automatically identify and mitigate spurious correlations through strategic data acquisition without requiring privileged information. In addition, these methods are designed for fixed datasets, whereas in active learning, the dataset is constructed iteratively through querying.

%% file: Appendix/E_Computational_Resources.tex
\section{Computational Resources}
\label{app:compute}

We report the computational resources used to reproduce the main experiments in Section~\ref{section:experiments}. All experiments were run on a single NVIDIA A100 GPU.

\paragraph{Per-run compute.}
Table~\ref{tab:compute_resources} reports the approximate wall-clock runtime for a single active-learning run on each dataset. Each run includes training across all active-learning rounds, from the initial labeled budget to the final labeled budget, including acquisition and model updates. CAML incurs additional computation relative to the standard active-learning baseline because it evaluates the cumulative meta-learning objective over queried batches from previous active-learning rounds.

\paragraph{Total compute.}
The main experiments evaluate the four datasets and two training procedures across four acquisition functions and three random seeds. The total compute required for the main results was approximately $(3.5+6+5+14+6+15+3.5+5)\times 4 \times 3 \approx 700$ GPU-hours. 

\begin{table}[h]
\centering
\caption{Approximate per-run computational cost. Runtime is reported as wall-clock hours for a single active-learning run on one NVIDIA A100 40GB GPU.}
\label{tab:compute_resources}
\begin{tabular}{lcc}
\toprule
Dataset & Standard Baseline & CAML \\
\midrule
Dominoes & 3.5h & 6h \\
Waterbirds & 5h & 14h \\
SpuCo & 6h & 15h \\
CivilComments & 3.5h & 5h \\
\bottomrule
\end{tabular}
\end{table}

%% file: Appendix/F_ExperimentResults.tex
\section{Additional Non-Spurious Benchmark}
\label{app:cifar100}

Our main empirical evaluation focuses on spurious-correlation benchmarks for three reasons. First, this setting directly extends the observation of ~\citet{tamkin2022active} that active-learning algorithms can be particularly effective in spurious-correlation settings by selecting examples that help distinguish the intended core correlation from the misleading spurious correlation. Second, these benchmarks provide a transparent way to diagnose the informativeness of queried batches: informative queries often correspond to minority-group examples that violate the dominant spurious correlation and are therefore necessary to disprove it. Third, they provide a direct measure of whether the learned inductive bias is moving in the intended direction. If CAML uses queried examples to refine the model's inductive bias toward core features and away from spurious features, this effect should be most clearly reflected in improved minority-group accuracy.

\begin{figure}[h]
    \centering
    \includegraphics[width=0.6\linewidth]{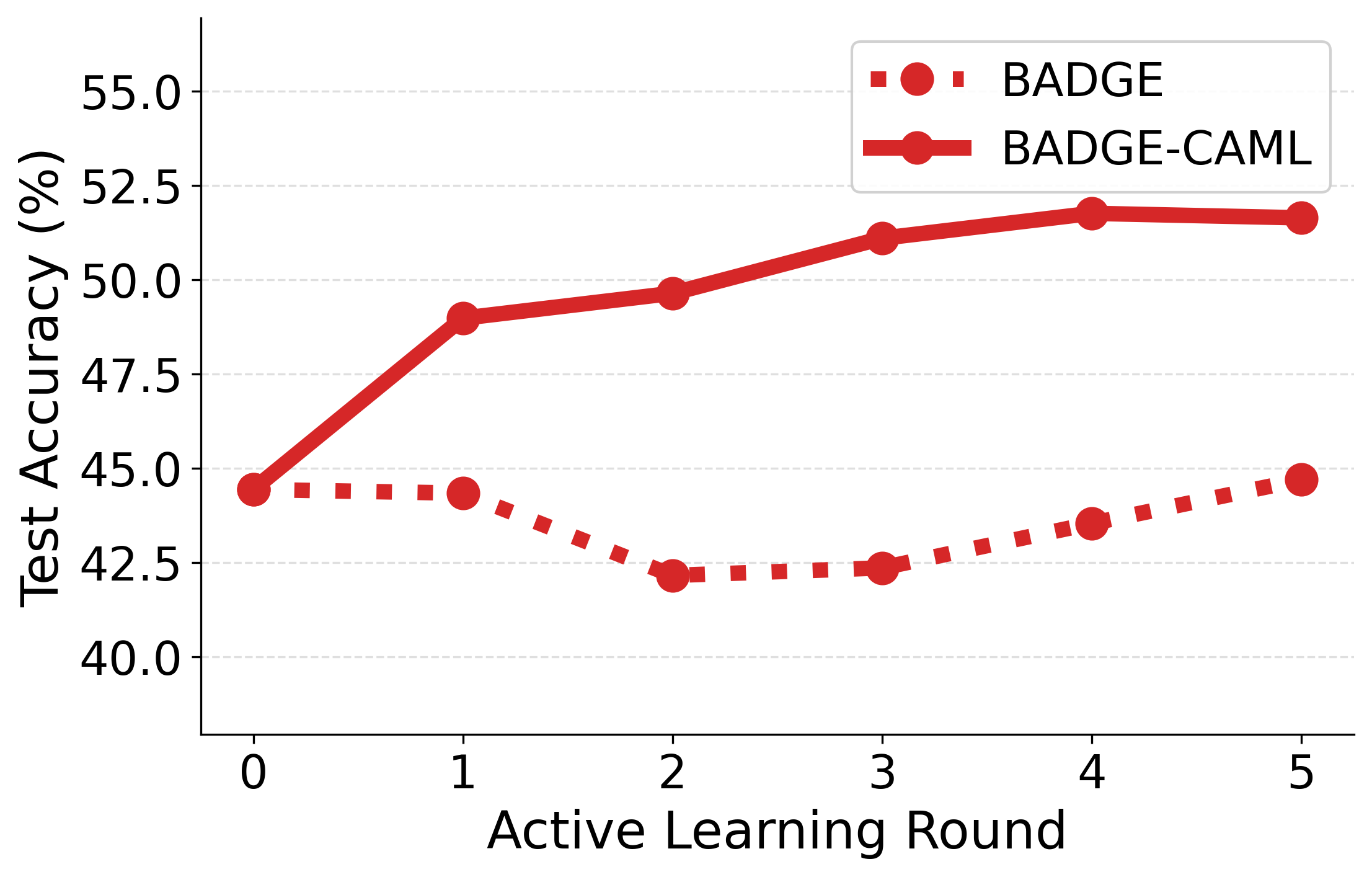}
    \caption{
    \textbf{Additional CIFAR100 results.}
    Standard test accuracy (\%) across active-learning rounds on CIFAR100 using BADGE acquisition. CAML improves standard test accuracy over the corresponding shuffle baseline, suggesting that query-based meta-learning can also provide benefits beyond explicitly spurious-correlation benchmarks.}
    \label{fig:cifar100}
\end{figure}

To provide an additional sanity check beyond explicitly spurious-correlation settings, we also evaluate CAML on CIFAR100, a standard image-classification benchmark without predefined spurious attributes or minority groups. Since CIFAR100 does not define worst-group performance, we report standard test accuracy. We use the same active-learning protocol as in the main experiments and compare the standard active-learning pipeline against CAML under BADGE acquisition. As shown in Figure~\ref{fig:cifar100}, CAML improves over the corresponding BADGE baseline across active-learning rounds, achieving a test-accuracy gain of $6.94\% \pm 1.34$. This result suggests that CAML can be extended beyond the spurious-correlation setting studied in the main paper. However, extensions to broader datasets and distribution-shift settings are outside the scope of this paper, which specifically studies robustness to spurious correlations, and remain an important direction for future work.
